\documentclass[10pt,twocolumn,letterpaper]{article}

\usepackage[pagenumbers]{cvpr} 

\usepackage{cite}
\usepackage{amsmath,amssymb,amsfonts}
\usepackage{algorithmic}
\usepackage{graphicx}
\usepackage{textcomp}

\usepackage{pifont}
\usepackage{booktabs}
\usepackage{algorithm}
\usepackage{textcomp}
\usepackage{multirow}
\usepackage{hhline, booktabs}
\usepackage{xcolor}
\usepackage{soul}   
\usepackage{colortbl}
\usepackage{float} 
\usepackage{caption} 
\usepackage{url}
\usepackage{enumerate}
\usepackage{tikz}

\usepackage{enumitem}
\usepackage{adjustbox}
\usepackage{comment}
\setlist[itemize]{left=0pt, itemsep=0pt, topsep=0pt}
\usepackage{scalefnt}

\definecolor{purple}{RGB}{166,128,184}
\definecolor{ForestGreen}{RGB}{34,139,34}
\definecolor{mycolor}{RGB}{178,200,222}

\definecolor{promptblue}{RGB}{15,117,188}
\definecolor{promptgreen}{RGB}{7,117,131}
\definecolor{promptpurple}{RGB}{166,128,184}
\definecolor{hlcolor}{RGB}{255,229,153} 

\sethlcolor{hlcolor} 

\usepackage{tikz}
\usepackage{lipsum}

\newlength{\RoundedBoxWidth}
\newsavebox{\GrayRoundedBox}
\newenvironment{GrayBox}[1][\dimexpr0.48\textwidth-4.5ex]%
   {\setlength{\RoundedBoxWidth}{\dimexpr#1}
    \begin{lrbox}{\GrayRoundedBox}
       \begin{minipage}{\RoundedBoxWidth}}%
   {   \end{minipage}
    \end{lrbox}
    \begin{center}
    \begin{tikzpicture}%
       \draw node[draw=black,fill=black!10,%
             inner sep=2ex,text width=\RoundedBoxWidth]%
             {\usebox{\GrayRoundedBox}};
    \end{tikzpicture}
    \end{center}}

\definecolor{cvprblue}{rgb}{0.21,0.49,0.74}
\usepackage[pagebackref,breaklinks,colorlinks,citecolor=cvprblue]{hyperref}

\def\paperID{*****} 
\def\confName{CVPR}
\def\confYear{2024}

\title{CBVLM: Training-free Explainable Concept-based Large Vision Language Models for Medical Image Classification}

\author{Cristiano Patrício\thanks{Equal contribution.\\Correspondence to: \texttt{cristiano.patricio@ubi.pt}} \textsuperscript{ 1,2,3}, Isabel Rio-Torto\textsuperscript{* 1,4}, Jaime S. Cardoso\textsuperscript{1,4}, Luís F. Teixeira\textsuperscript{1,4}, João C. Neves\textsuperscript{2,3}
\\
\small\textsuperscript{1}INESC TEC
\quad\textsuperscript{2}NOVA LINCS
\quad\textsuperscript{3}Universidade da Beira Interior
\quad\textsuperscript{4}Universidade do Porto
}

\begin{document}
\maketitle

\begin{abstract}
The main challenges limiting the adoption of deep learning-based solutions in medical workflows are the availability of annotated data and the lack of interpretability of such systems. Concept Bottleneck Models (CBMs) tackle the latter by constraining the {model output} on a set of predefined and human-interpretable concepts. However, the increased interpretability achieved through these concept-based explanations implies a higher annotation burden. Moreover, if a new concept needs to be added, the whole system needs to be retrained. Inspired by the remarkable performance shown by Large Vision-Language Models (LVLMs) in few-shot settings, we propose a simple, yet effective, methodology, \emph{CBVLM}, which tackles both of the aforementioned challenges. First, for each concept, we prompt the LVLM to answer if the concept is present in the input image. Then, we ask the LVLM to classify the image based on the previous concept predictions. Moreover, in both stages, we incorporate a retrieval module responsible for selecting the best examples for in-context learning. By grounding the final diagnosis on the predicted concepts, we ensure explainability, and by leveraging the few-shot capabilities of LVLMs, we drastically lower the annotation cost. We validate our approach with extensive experiments across four medical datasets and twelve LVLMs (both generic and medical) and show that \emph{CBVLM} consistently outperforms CBMs and task-specific supervised methods without requiring any training and using just a few annotated examples. More information on our project page: \url{https://cristianopatricio.github.io/CBVLM/}.
\end{abstract}

\begin{figure}[t]
\centering
\includegraphics[width=\linewidth]{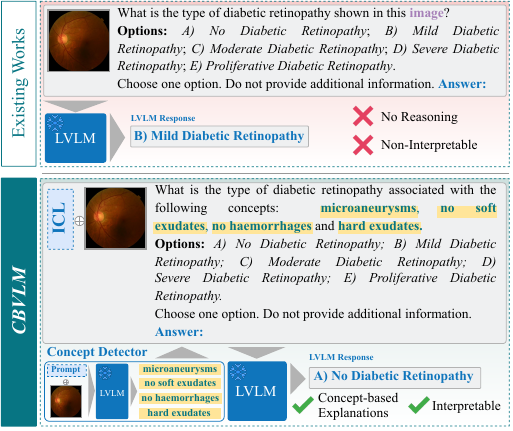}
\vspace{-5mm}
\caption{\textbf{Comparison between the proposed methodology (\emph{CBVLM}) with existing methods.} Unlike traditional approaches, our methodology grounds the final diagnosis on a set of clinical concepts predicted by the LVLM itself (\textit{cf.} \textit{Concept Detector}), thus increasing classification performance, while being interpretable.}
\label{fig:figure_intro}
\end{figure}

\section{Introduction}
\label{sec:introduction}

The field of medical imaging analysis has witnessed significant advancements through the use of supervised deep learning methods. However, their performance greatly depends on the availability and quality of training samples, which are particularly costly to obtain, as clinician expertise is required. The recent rise of Large Vision-Language Models (LVLMs) has also made its way into the medical field, of which Med-LVLMs such as~\cite{moor2023med, li2024llava, chen2024chexagent, zhou2024pre} are prominent examples, where pre-trained LVLMs are fine-tuned using medical data. However, despite the improved performance, these Med-LVLMs require even more annotated data. Consequently, a different research direction has emerged, exploring the potential of applying LVLMs to specific downstream tasks with minimal or no supervision, by leveraging their few-shot capabilities~\cite{han2023multimodal, van2024large,ferber2024context}, particularly the use of in-context learning (ICL) for improving model performance through a short number of examples in the prompt. 
In addition to the need for annotated data, another major challenge with deep learning-based medical imaging is the lack of interpretability, which is of the utmost importance in high-stakes decision-making scenarios~\cite{rudin2019stop}. One way of introducing interpretability into such frameworks is via concept-based explanations. Concept Bottleneck Models (CBMs) provide this type of explanations by introducing a layer, the ``bottleneck'', that predicts the presence or absence of a predefined set of human-interpretable concepts. The final classification prediction is then based on these concepts. Nevertheless, CBMs still require training and annotated data; in fact, the amount of annotated data actually increases since concept annotations are needed on top of the classification labels.
We propose combining the best of both worlds: the interpretability of CBMs and the few-shot abilities of LVLMs to decrease the annotation cost. Our methodology, \emph{CBVLM}, encompasses a two-stage approach, where, first, the LVLM is prompted to predict the existence/absence of a set of concepts in an image, and, in a second stage, the predicted concepts are incorporated into the prompt such that the LVLM provides a final diagnosis grounded on these predicted concepts. Figure~\ref{fig:figure_intro} represents this methodology compared to the common approach where an LVLM is simply prompted to perform classification. 
Moreover, contrary to CBMs, which require retraining the entire model to incorporate a new concept, \emph{CBVLM} can easily incorporate additional concepts, requiring only two inference steps.
We validate \emph{CBVLM} on four medical datasets encompassing different imaging modalities and using twelve LVLMs, and show that \emph{CBVLM} achieves comparable performance or even surpasses task-specific supervised methods, doing so without training, with few annotated examples, and without sacrificing interpretability. Although, individually, the techniques used in \emph{CBVLM} (e.g., ICL) are not novel, this is, to the best of our knowledge, the first work combining them in a novel methodology exploring the capability of LVLMs to provide concept-based explanations for medical image classification.

\begin{figure*}[ht]
\centering
\includegraphics[width=\linewidth]
{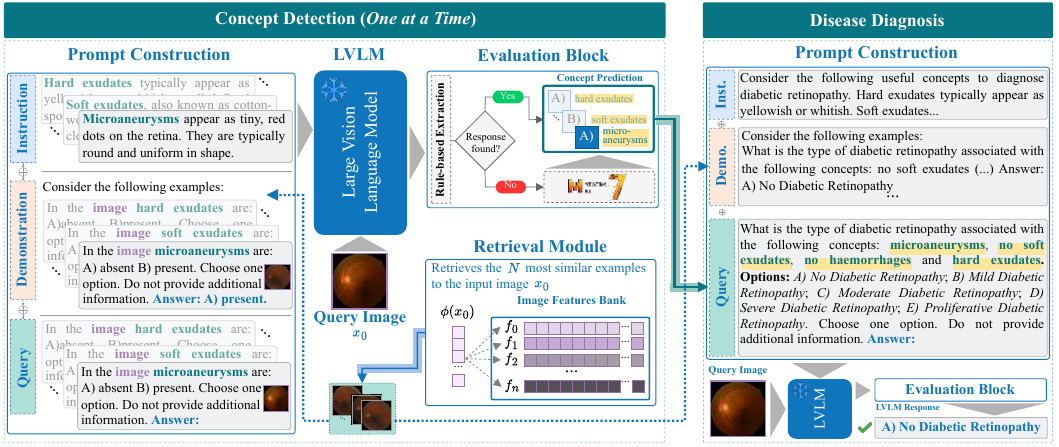}
 \caption{\textbf{Overview of \emph{CBVLM}.} Our methodology is organized into two key stages: 1) The \textit{\underline{Concept Detection}} stage, where the LVLM predicts the individual presence of each predefined clinical concept in the query image. This is achieved using a custom prompt that supports both zero- and few-shot settings. In the latter, we include a set of demonstration examples (middle block of \textit{Prompt Construction}) chosen by the \textit{Retrieval Module}, responsible for selecting the $N$ most similar examples to the input image. To evaluate the LVLM answer, we employ an \textit{Evaluation Block} which first tries to extract the desired LVLM response using a rule-based formulation. If this fails, we adopt an auxiliary LLM to extract the desired response. 2) In the \textit{\underline{Disease Diagnosis}} stage, the final diagnosis is generated by the LVLM based on the clinical concepts predicted in the first stage, which are directly incorporated in the query (highlighted in yellow). This approach ensures that the diagnosis is grounded on the identified clinical concepts, enhancing the interpretability and transparency of the LVLM’s response. In this second stage, the \textit{Retrieval Module} is also used to select the $N$ most similar demonstrations.}
\label{fig:pipeline}
\end{figure*}

\section{Related Work}
\label{sec:related_work}

\paragraph{\textbf{Large Vision-Language Models in the Medical Domain}}
The emergent capabilities of Large Language Models (LLMs) have led to the development of LLMs able to process visual inputs, i.e. LVLMs. These models can reason about images in a zero- or few-shot manner (e.g., LLaVA~\cite{liu2023llava}, GPT-4V~\cite{OpenAI_GPT4_2023}), allowing them to tackle downstream tasks without fine-tuning. General-purpose LVLMs have recently been adapted for the medical domain through training on extensive medical datasets, coining the term Med-LVLMs. The generalization capability of these models has prompted researchers to assess their performance in multiple medical data. Royer et al.~\cite{royer2024multimedeval} recently released an open-source toolkit for fair and reproducible evaluation of Med-LVLMs in tasks like classification and report generation. Han et al.~\cite{han2023multimodal} evaluated the performance of a general LVLM on the NEJM Image Challenge, as well as the impact of fine-tuning the LVLM. In~\cite{wang2024asclepius}, Med-LVLMs were evaluated in different specialities and compared with medical experts, with results comparable to the performance of junior doctors. In~\cite{ferber2024context}, the authors benchmarked the performance of GPT-4V against fully supervised image classifiers and found that their performance is comparable. In a different line of research, Xia et al.~\cite{xia2024cares} evaluated the trustworthiness and fairness of Med-LVLMs and found that models often display factual inaccuracies and can be unfair across different demographic groups.
\emph{CBVLM} distinguishes itself from previous studies as it is not limited to prompting the LVLM to provide a diagnosis, but it is, to the best of our knowledge, the first to generate concept-based explanations on which the LVLM response is grounded prior to diagnosis prediction, thus enhancing the performance and transparency of the decision-making process.

\paragraph{\textbf{Concept Bottleneck Models}}
CBMs~\cite{koh2020concept} are based on an encoder-decoder paradigm, where the encoder predicts a set of human-specified concepts from the input image, while the decoder leverages these predicted concepts to generate the final predictions. Due to their inherent interpretability, these models have attracted particular interest in medical image classification. Several CBMs have been proposed for a wide variety of medical tasks (e.g., melanoma diagnosis~\cite{patricio2023coherent, bie2024mica, patricio2024towards}, knee osteoarthritis~\cite{koh2020concept} and chest X-ray classification~\cite{ghosh2023bridging}). Nevertheless, CBMs have well-known limitations: (i) they require concept annotations (image-level or class-level), (ii) they tend to have lower performance compared to traditional (black-box) supervised methods, (iii) they struggle to predict concepts accurately, and (iv) incorporating new concepts implies retraining the whole CBM. These challenges are particularly significant in medical imaging, where data availability is typically limited due to the expertise required for annotations. Consequently, LVLMs represent a promising alternative to traditional CBMs, as they are able to generalize to unseen domains with minimal or no supervision. Moreover, incorporating additional concepts becomes trivial and does not require fine-tuning, as happened with traditional CBMs.

\section{Methodology}
\label{sec:methodology}

Given the task of predicting a target $y \in \mathbb{R}$ from input $x \in \mathbb{R}^d$, let $\mathcal{D} = \{(x^{(i)}, y^{(i)}, c^{(i)})\}_{i=1}^{n}$ be a batch of training samples where {$c \in \mathbb{R}^{\{0,1\}^l}$} is a vector of $l$ concepts. Traditional (black-box) models learn a function $h: \mathbb{R}^d \rightarrow \mathbb{R}$ that directly predicts the target $y \in \mathbb{R} $ from the input $x \in \mathbb{R}^d$. However, CBMs first map the input $x$ into a set of interpretable concepts $c$ (``bottleneck'') by learning $g: \mathbb{R}^d \rightarrow \mathbb{R}^l $, and use these concepts to predict the target $y$ using $f: \mathbb{R}^l \rightarrow \mathbb{R}$. Therefore, the prediction $\hat{y} = f(g(x))$ is entirely based on the predicted concepts $\hat{c} = g(x)$.

In order to generate concept-based explanations with LVLMs, we follow the CBM paradigm and introduce a two-stage methodology, as outlined in Figure~\ref{fig:pipeline} and detailed in the following subsections.

\subsection{Concept Detection}

The first stage involves predicting the $l$ concepts present in a given input image. For each predefined clinical concept, we start by providing the LVLM with a short description of that concept and then prompt it to answer if that concept is shown in the image (see Figure~\ref{fig:pipeline} on the left). Thus, concept prediction is given by $p_{LVLM}(t_{ans} | t_I^1, ..., t_I^m, t_P^1, ..., t_P^{k})$, where $t_{ans}$ corresponds to the first generated token, $t_I^1, ..., t_I^m$ correspond to the input image tokens, and $t_P^1, ..., t_P^k$ correspond to the tokens of the prompt (description of the concept and the question). The structure of the (zero-shot) prompt for the prediction of a single concept is as follows:

\begin{GrayBox}[0.95\columnwidth]
\scalefont{0.75}

\{A brief description of the \{\textcolor{promptgreen}{concept}\}\}.

In the \textcolor{promptpurple}{\textless image\textgreater}, the \{\textcolor{promptgreen}{concept}\} is:

 A) present
 
 B) absent
 
Choose one option. Do not provide additional information. \textcolor{promptblue}{Answer:}
\end{GrayBox}

The descriptions for each clinical concept, as presented in the \textit{Instruction} section shown in Figure \ref{fig:pipeline}, were generated by ChatGPT~\cite{OpenAI_ChatGPT} with the following prompt: "\textit{According to published literature in \{dermatology, radiology, retinal lesions\}, which phrases best describe a \{skin, X-ray, eye fundus\} image containing {\{concept\}}?"}.

Examples of prompts for different concepts related to the classification of diabetic retinopathy are shown in Figure~\ref{fig:pipeline} on the left. The full prompts for all datasets can be found in the Supplementary Material.

\subsection{Disease Diagnosis}
The second stage involves prompting the LVLM to answer if a certain disease (i.e. class) is present in the image. The concepts predicted in the previous stage are appended to the question, such that when providing an answer, the LVLM not only considers the image and the question, but also takes into account the concepts. Thus, the disease diagnosis is given by $p_{LVLM}(t_{ans} | t_I^1, ..., t_I^m, t_P^1, ..., t_P^k, t_C^1, ..., t_C^c)$, where $t_C^1, ..., t_C^c$ are the tokens representing the concepts. This approach ensures that the diagnosis is grounded on the identified clinical concepts, enhancing the interpretability and transparency of the model’s response. The structure of the (zero-shot) prompt for diagnosis classification is as follows:

\begin{GrayBox}[0.95\columnwidth]
\scalefont{0.75}

\{A brief description of \{\textcolor{promptgreen}{concept $0$}\}\}.

(...)

\{A brief description of \{\textcolor{promptgreen}{concept $l-1$}\}\}.

What is the diagnosis that is associated with the following concepts: \{\textcolor{promptgreen}{\hl{predicted concepts}}\}

\textbf{Options:}

 A) class 0
 
 B) class 1

 (...)

Choose one option. Do not provide additional information. \textcolor{promptblue}{Answer:}
\end{GrayBox}

An example of the prompt designed for the classification of diabetic retinopathy is shown in Figure~\ref{fig:pipeline} on the right. The full prompts for all datasets can be found in the Supplementary Material.

\subsection{Few-shot Prompting}

A key factor contributing to the appeal of LLMs, and consequently LVLMs, is their ability to perform ICL.
ICL, or few-shot learning/prompting, is a prompt engineering technique in which the model is provided with examples, known as demonstrations, consisting of: i) questions, and ii) their corresponding answers. It has been shown to significantly improve the performance of both LLMs~\cite{gpt3} and LVLMs~\cite{alayrac2022flamingo}. Therefore, our methodology incorporates ICL in both stages, as shown in Figure~\ref{fig:pipeline}.

\subsection{Demonstration Example Retrieval}
Although the demonstrations for ICL could be chosen randomly, Zhou et al.~\cite{zhou2024adapting} have shown that this selection policy can decrease the model performance when compared to the 0-shot scenario (when no examples are included in the prompt). Thus, following other works~\cite{alayrac2022flamingo, awadalla2023openflamingo, yang2022empirical, chen2024understanding}, we introduce a Retrieval Module to select the demonstrations based on their similarity to the query image $x_0$, a method known as Retrieval-based In-Context Examples Selection (RICES). Given the feature vector $\phi(x_0)$ of the query image obtained from an arbitrary vision encoder, we select the $N$ most similar image feature vectors from the training set $\mathcal{D}$ as the few-shot demonstrations to include in the prompt. For each different type of prompt, the question is repeated for every demonstration. The ground-truth answer to each demonstration question is also provided; depending on the stage of our methodology, this can be the absence/presence of a given concept (1st stage), or the disease diagnosis, as well as the absence/presence of all concepts (2nd stage).

\label{subsec:icl_demo_selection}

\subsection{Answer Extraction}

As LVLMs produce their outputs in natural language (i.e. it is open-ended), it is necessary to process the generated text to compare it with the ground-truth concepts or the ground-truth class labels. To simplify this process and limit the space of possible responses, all questions are posed as multiple choice and the LVLM is instructed to answer by choosing one of the options and not providing additional information (see the example prompts in Figure~\ref{fig:pipeline}). 

One way of extracting the LVLM answer would be to select the option with the higher first token log probability, i.e. if token B is the one with the highest probability, then this would mean that the LVLM had chosen option B. However,~\cite{wang2024-answerc} have shown that there is a mismatch between the first token probabilities and the actual LVLM answer. Thus, we use a Regular Expression (RegEx) to directly find the pattern ``option\_letter)'', either in lower or upper case. We choose to provide the options as ``option\_letter'' + ``)'' instead of ``option\_letter'' + ``.'', as it would be less probable to find the pattern ``option\_letter)'' in the middle of the LVLM response. If none of the predefined options is found in the LVLM answer (e.g., when the LVLM answers the question but in a more verbose fashion), we follow the approach of previous works~\cite{chen2024chexagent} and employ an auxiliary LLM\footnote{\texttt{Mistral-7B-Instruct-v0.3}} to extract the information from the LVLM answer and match it to the predefined list of options. If both approaches fail (e.g., when the LVLM response is just a description of the input image without any actual answer to the question), we mark that instance as unknown and report separately the percentage of instances for which this occurred.

\section{Experimental Setup}
\label{sec:experiments}
In this section, we outline our experimental setup with information about the adopted LVLMs, datasets, and evaluation metrics, as well as the implementation details.

\subsection{Datasets}
\label{subsec:datasets} 
We conduct experiments on four publicly available datasets, encompassing dermatology (\textbf{Derm7pt}~\cite{DERM7PT} and \textbf{SkinCon}~\cite{daneshjou2022skincon}), radiology (\textbf{CORDA}~\cite{cordadataset}) and eye fundus imaging (\textbf{DDR}~\cite{li2019diagnostic}) modalities. The selection of these datasets is constrained by the availability of datasets with annotated clinical features, hereinafter referred to as ``\textit{concepts}''. Unless otherwise stated, we use the official train-val-test splits of the datasets.

For the \textbf{Derm7pt}~\cite{DERM7PT} dataset, with a total of 827 images, we consider only dermoscopic images of the ``nevus'' and ``melanoma'' classes, jointly with the 5 annotated dermoscopic attributes, namely Absent/Typical/Atypical Pigment Network, Absent/Regular/Irregular Streaks, Absent/Regular/Irre\-gular Dots and Globules, Blue-Whitish Veils, and Regression Structures.
The \textbf{SkinCon}~\cite{daneshjou2022skincon} dataset contains 3,230 images from the Fitzpatrick 17k dataset~\cite{groh2021evaluating} annotated with 48 clinical concepts. Following previous works, we use the binary labels denoting skin malignancy for the target task and randomly split the dataset into train/val/test with the proportion of 70\%, 15\%, and 15\%, respectively. Additionally, only 22 out of the 48 concepts are used, as the other 26 concepts appear in less than 50 images. 
From the \textbf{CORDA}~\cite{cordadataset} dataset, a COVID-19 diagnosis dataset, we use its 1601 chest X-rays. Inspired by the work of Barbano et al.~\cite{barbano2022two}, we use a pretrained model on the CheXpert~\cite{irvin2019chexpert} dataset to annotate the samples with a binary label to denote the presence or absence of 14 radiological observations. {Of those 14 concepts, we exclude the ``Fracture'' and ``Support Devices'' concepts, as these are not directly related to the presence or absence of the target class. The “No Finding” concept is implicitly included in the remaining 11 concepts given that if all 11 concepts are 0 it represents a healthy case without any findings.}
The \textbf{DDR}~\cite{li2019diagnostic} dataset comprises 13,673 fundus images divided into six Diabetic Retinopathy (DR) levels. In our experiments, we focus on the five-class classification task for DR grading, discarding the images belonging to the ungradable class and considering a subset of 757 images annotated with 4 types of lesions correlated with DR, which we use as concepts. We augment this subset with 743 ``No DR'' images, resulting in a dataset of 1,500 samples.

Table \ref{tab:dataset_statistics} provides a detailed summary of the statistics for each of the considered datasets. All results (\textit{cf.} Section~\ref{sec:results}) are reported on the entire test set of each dataset. 
\begin{table}[h]
\caption{Detailed statistics for the datasets used.}
\label{tab:dataset_statistics}
\begin{center}
\resizebox{0.8\linewidth}{!}{%
\begin{tabular}{llcccc}
\toprule
\textbf{Dataset}& 
\textbf{Image Type \& Task}&
\textbf{Classes}&
\textbf{Concepts}&
\textbf{Train}& 
\textbf{Test}\\
\midrule
\multirow{2}{*}{Derm7pt~\cite{DERM7PT}} & Dermoscopic & \multirow{2}{*}{2} & \multirow{2}{*}{5} & \multirow{2}{*}{346} & \multirow{2}{*}{320} \\
&  Melanoma classification &&&&\\
\midrule
\multirow{2}{*}{SkinCon~\cite{daneshjou2022skincon}} & Dermoscopic & \multirow{2}{*}{2} & \multirow{2}{*}{22} & \multirow{2}{*}{2582} & \multirow{2}{*}{554} \\
& Malignancy classification  &&&& \\
\midrule
\multirow{2}{*}{CORDA~\cite{cordadataset}} & X-ray & \multirow{2}{*}{2} & \multirow{2}{*}{{11}} & \multirow{2}{*}{967} & \multirow{2}{*}{392} \\
& Covid-19 classification &&&&\\
\midrule
\multirow{2}{*}{DDR~\cite{li2019diagnostic}} & Fundus & \multirow{2}{*}{5} & \multirow{2}{*}{4} & \multirow{2}{*}{817} & \multirow{2}{*}{404} \\
& DR classification  &&&&\\
\bottomrule
\end{tabular}%
}
\end{center}
\end{table}

\subsection{LVLMs}
\label{subsec:methods}

In our experiments, we assess the performance of open-source LVLMs that can process both text and multiple images. Specifically, we focus on models trained on general-domain data, such as OpenFlamingo~\cite{awadalla2023openflamingo}, Idefics3~\cite{idefics3}, VILA~\cite{lin2024vila}, LLaVA-OneVision~\cite{li2024llavaov}, Qwen2-VL~\cite{Qwen-VL}, MiniCPM-V~\cite{yao2024minicpm}, InternVL 2.5~\cite{chen2025internvl25} and mPLUG-Owl3~\cite{ye2024mplugowl3longimagesequenceunderstanding}. Additionally, we consider models specialized for the medical domain, including Med-Flamingo~\cite{moor2023med}, LLaVA-Med~\cite{li2024llava}, CheXagent~\cite{chen2024chexagent} and SkinGPT-4~\cite{zhou2024pre}. We use their corresponding checkpoints from the HuggingFace library (the full list of checkpoints can be found in the Supplementary Material).

\subsection{Baselines}
We compare the performance of \emph{CBVLM} against several baselines, including: 

i) \textbf{standard CBM}~\cite{koh2020concept} trained on each of the four datasets considered using the official implementation. It is worth noting that, for training the CBM on the Derm7pt dataset, we considered 11 clinical concepts. Specifically, 3 out of the 5 annotated concepts have three options each (e.g., Pigment Network can be Absent, Regular, or Irregular), as described in \ref{subsec:datasets}. To address this, and given that the concept layer predicts the presence or absence of each concept, we treated each option as an independent concept. For instance, for Pigment Network, we defined the concepts: Absent Pigment Network, Regular Pigment Network, and Irregular Pigment Network;

ii) \textbf{CLAT}~\cite{CLAT} - a concept-based framework specifically designed for retinal disease diagnosis. For DDR, 
the pre-computed textual embeddings of the clinical concepts provided in the official CLAT repository were used. However, since the original embeddings were obtained from a retinal foundation model~\cite{silva2025foundation}, for the other datasets, we replaced it with MedImageInsight~\cite{codella2024medimageinsight}, an embedding model for medical imaging trained on a mixture of X-ray, computed tomography, dermatology, and pathology datasets. For Derm7pt, we follow the same strategy described above, i.e. we transformed the 5 concepts into 11 independent concepts;

iii) \textbf{supervised black-box models}, namely ImageNet-pretrained Res\-Net50~\cite{he2016deep} and ViT Base~\cite{dosovitskiy2020image} fine-tuned on the disease diagnosis task; 

iv) \textbf{task-specific black-box models}, namely~\cite{patricio2024towards} for the Derm7pt data\-set,~\cite{hou2024concept} for the SkinCon dataset,~\cite{barbano2024ai} for the CORDA dataset, and~\cite{madarapu2024multi} for the DDR dataset, reporting classification performance on the respective dataset.

\subsection{Retrieval Module}
For the retrieval strategy used to perform ICL, image similarity is determined using cosine similarity between features, which are extracted from two distinct encoders, namely MedImageInsight~\cite{codella2024medimageinsight} and the (vision) encoder of the corresponding LVLM.

\subsection{Evaluation Metrics}
\label{subsec:metrics}

For both concept prediction and disease diagnosis, we report Balanced Accuracy (BACC) and F1-score, as these metrics are tailored to deal with imbalanced datasets, which is the case in medical image classification. For concept prediction, we measure the BACC and F1-score averaged over each concept.

\begin{figure*}[!t]
    \centering
    \begin{tabular}{c}
        \vspace{-3mm}
        \begin{tikzpicture}
            \node at (0, 0) {\includegraphics[width=0.9\textwidth]{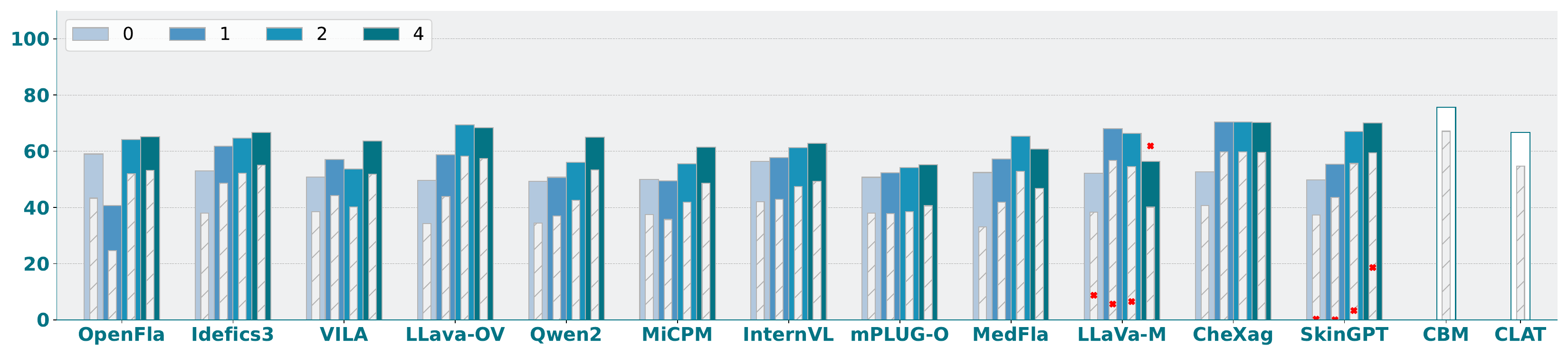}};
            \node[font=\footnotesize\sffamily\bfseries, rotate=90, text=black] at (-0.46\textwidth, 0) {Derm7pt};
            \node[font=\footnotesize\sffamily\bfseries, text=gray] at (-0.165\textwidth, 2) {Generic};
            \node[font=\footnotesize\sffamily\bfseries, text=gray] at (0.23\textwidth, 2) {Medical};
            \draw[dashed, thick, gray] (0.099\textwidth, -1.35) -- (0.099\textwidth, 1.645);
            \draw[dashed, thick, gray] (0.355\textwidth, -1.35) -- (0.355\textwidth, 1.645);
        \end{tikzpicture} \\
        \vspace{-3mm}
        \begin{tikzpicture}
            \node at (0, 0) {\includegraphics[width=0.9\textwidth]{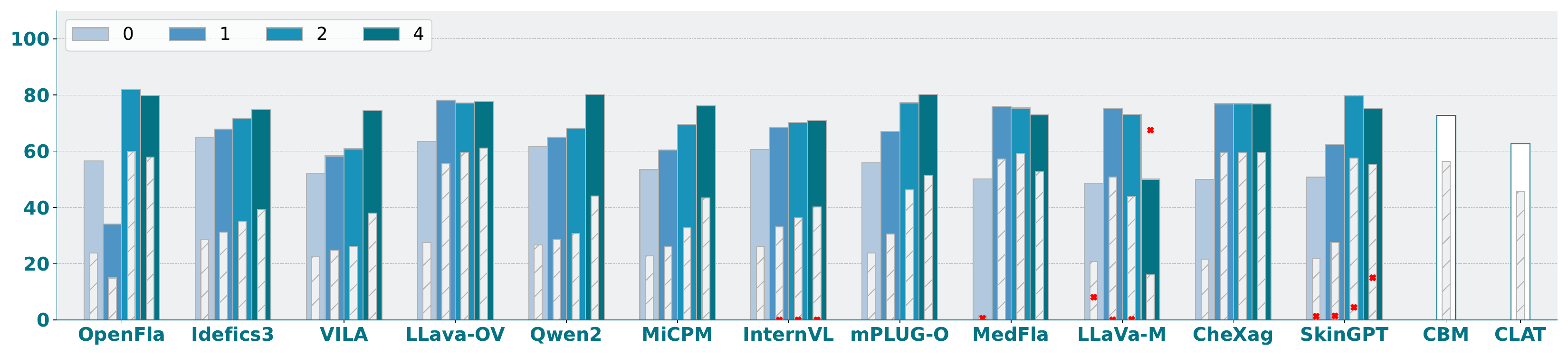}};
            \node[font=\footnotesize\sffamily\bfseries, rotate=90, text=black] at (-0.46\textwidth, 0) {SkinCon};
            \draw[dashed, thick, gray] (0.099\textwidth, -1.35) -- (0.099\textwidth, 1.645);
            \draw[dashed, thick, gray] (0.355\textwidth, -1.35) -- (0.355\textwidth, 1.645);
        \end{tikzpicture} \\
        \vspace{-3mm}
        \begin{tikzpicture}
            \node at (0, 0) {\includegraphics[width=0.9\textwidth]{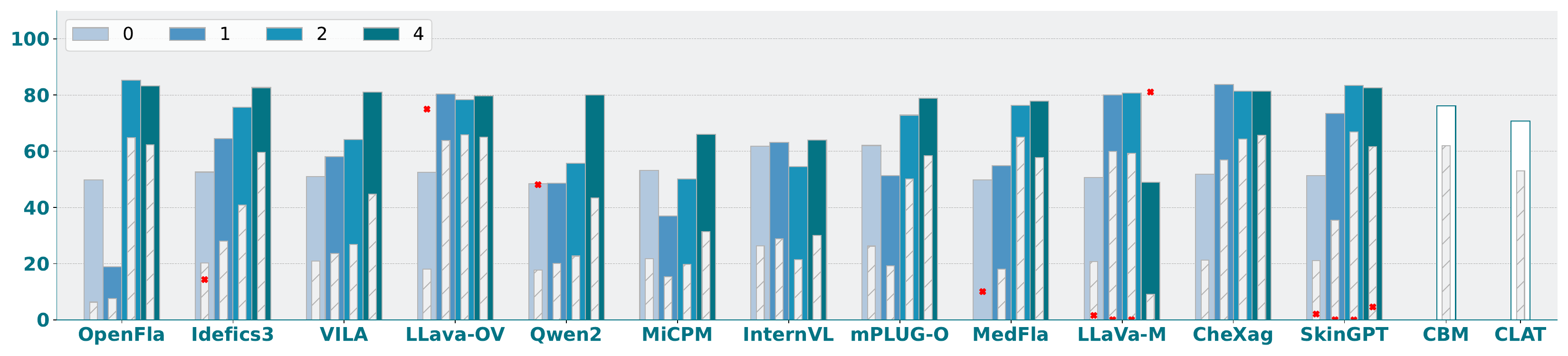}};
            \node[font=\footnotesize\sffamily\bfseries, rotate=90, text=black] at (-0.46\textwidth, 0) {CORDA};
            \draw[dashed, thick, gray] (0.099\textwidth, -1.35) -- (0.099\textwidth, 1.645);
            \draw[dashed, thick, gray] (0.355\textwidth, -1.35) -- (0.355\textwidth, 1.645);
        \end{tikzpicture} \\
        \begin{tikzpicture}
            \node at (0, 0) {\includegraphics[width=0.9\textwidth]{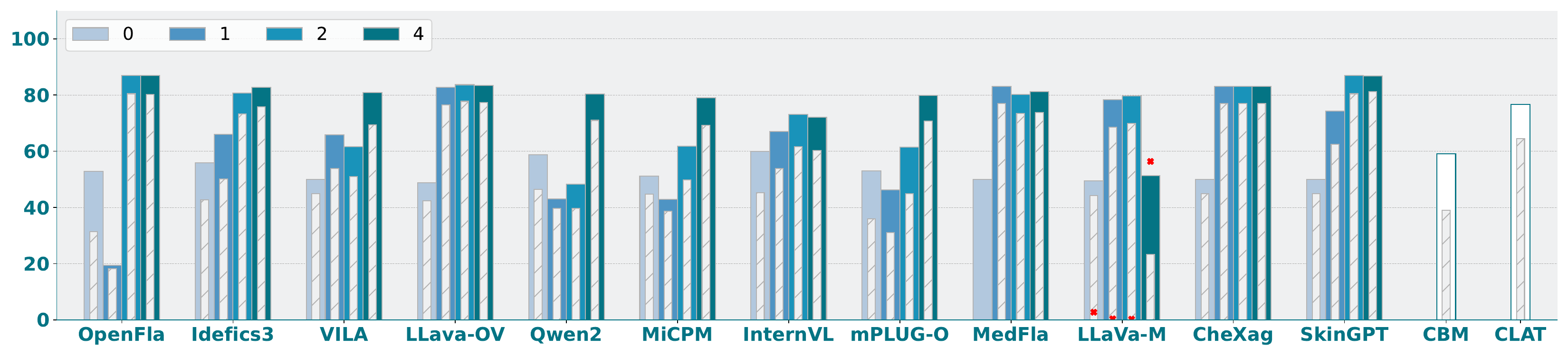}};
            \node[font=\footnotesize\sffamily\bfseries, rotate=90, text=black] at (-0.46\textwidth, 0) {DDR};
            \draw[dashed, thick, gray] (0.099\textwidth, -1.35) -- (0.099\textwidth, 1.645);
            \draw[dashed, thick, gray] (0.355\textwidth, -1.35) -- (0.355\textwidth, 1.645);
        \end{tikzpicture}
    \end{tabular}
    \vspace{-3mm}
    \caption{{\textbf{Concept detection performance of LVLMs across different $n$-shot settings}. Each bar corresponds to an $n$-shot scenario ($n = \{0,1,2,4\}$). Filled colored bars denote BACC, whereas the hatched bars indicate F1-scores. Red crosses indicate the percentage of unknowns, i.e. the proportion of samples whose LVLMs' responses do not contain sufficient information to answer the posed question.}}
    \label{fig:results_concept_detection}
\end{figure*}

\section{Results and Analysis}
\label{sec:results}
\begin{figure*}[!t]
    \centering
    \begin{tabular}{c}
        \vspace{-3mm}
        \begin{tikzpicture}
            \node at (0, 0) {\includegraphics[width=0.9\textwidth]{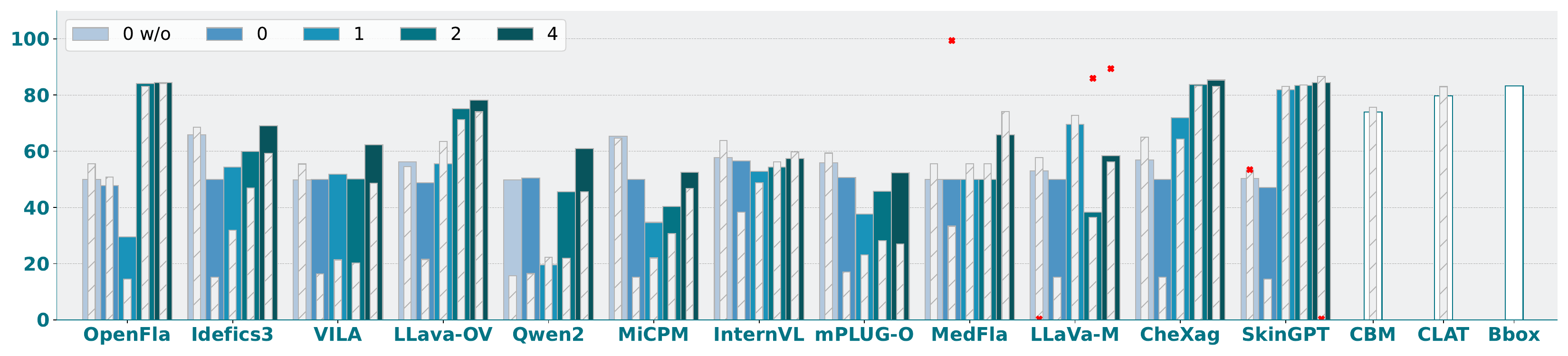}}; 
            \node[font=\footnotesize\sffamily\bfseries, rotate=90, text=black] at (-0.46\textwidth, 0) {Derm7pt};
            \node[font=\footnotesize\sffamily\bfseries, text=gray] at (-0.18\textwidth, 2) {Generic};
            \node[font=\footnotesize\sffamily\bfseries, text=gray] at (0.20\textwidth, 2) {Medical};
            \draw[dashed, thick, gray] (0.076\textwidth, -1.35) -- (0.076\textwidth, 1.645);
            \draw[dashed, thick, gray] (0.321\textwidth, -1.35) -- (0.321\textwidth, 1.645);
        \end{tikzpicture} \\
        \vspace{-3mm}
        \begin{tikzpicture}
            \node at (0, 0) {\includegraphics[width=0.9\textwidth]{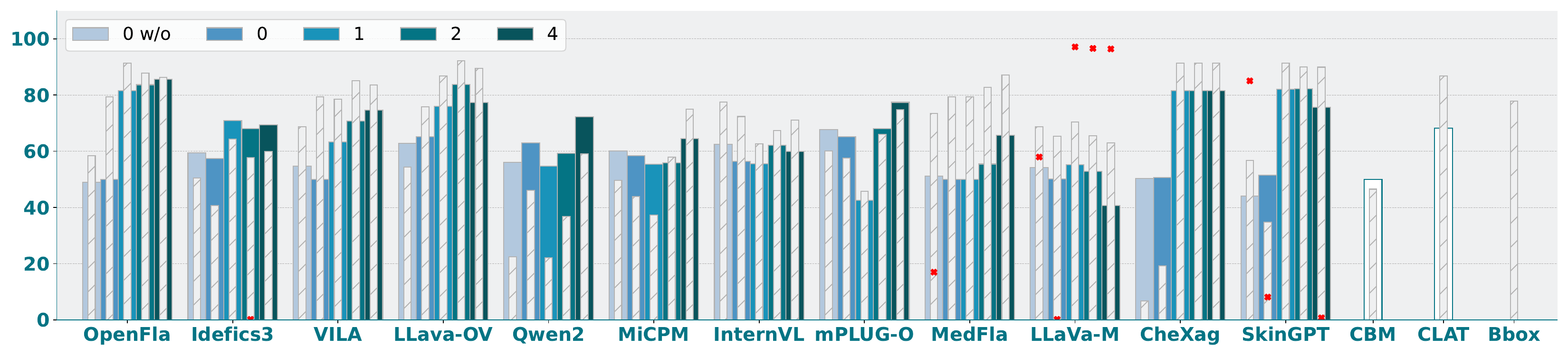}}; 
            \node[font=\footnotesize\sffamily\bfseries, rotate=90, text=black] at (-0.46\textwidth, 0) {SkinCon};
            \draw[dashed, thick, gray] (0.076\textwidth, -1.35) -- (0.076\textwidth, 1.645);
            \draw[dashed, thick, gray] (0.321\textwidth, -1.35) -- (0.321\textwidth, 1.645);
        \end{tikzpicture} \\
        \vspace{-3mm}
        \begin{tikzpicture}
            \node at (0, 0) {\includegraphics[width=0.9\textwidth]{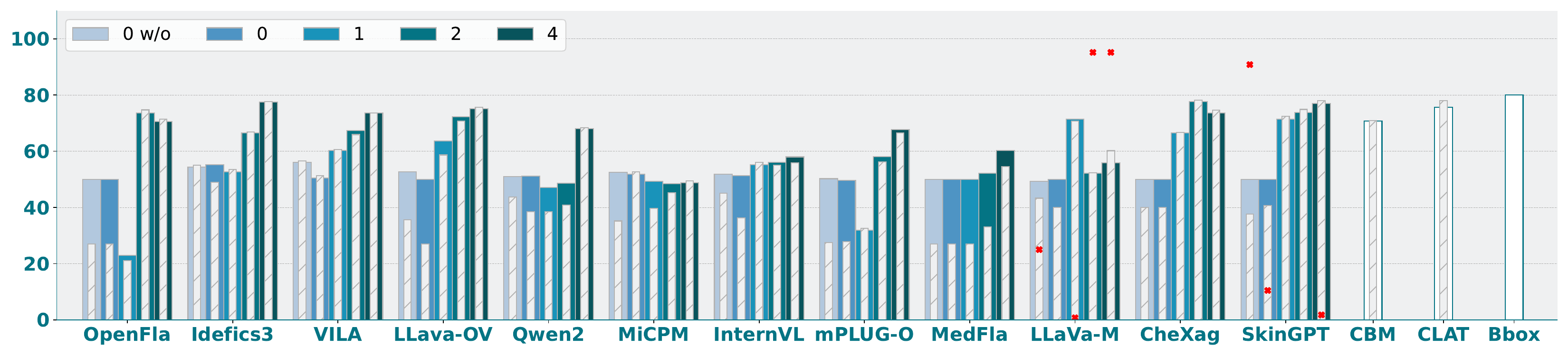}};
            \node[font=\footnotesize\sffamily\bfseries, rotate=90, text=black] at (-0.46\textwidth, 0) {CORDA};
            \draw[dashed, thick, gray] (0.076\textwidth, -1.35) -- (0.076\textwidth, 1.645);
            \draw[dashed, thick, gray] (0.321\textwidth, -1.35) -- (0.321\textwidth, 1.645);
        \end{tikzpicture} \\
        \begin{tikzpicture}
            \node at (0, 0) {\includegraphics[width=0.9\textwidth]{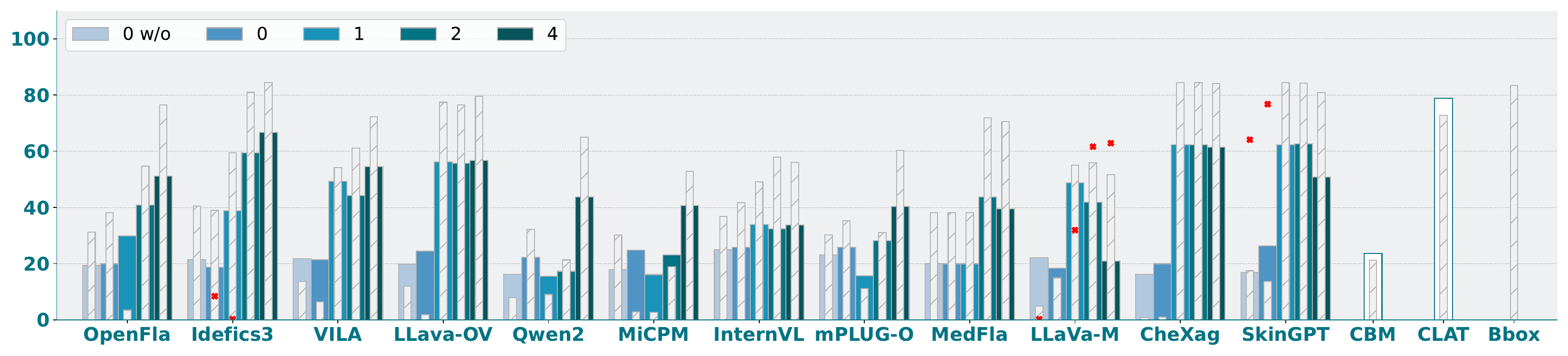}};
            \node[font=\footnotesize\sffamily\bfseries, rotate=90, text=black] at (-0.46\textwidth, 0) {DDR};
            \draw[dashed, thick, gray] (0.076\textwidth, -1.35) -- (0.076\textwidth, 1.645);
            \draw[dashed, thick, gray] (0.321\textwidth, -1.35) -- (0.321\textwidth, 1.645);
        \end{tikzpicture}
    \end{tabular}
    \vspace{-3mm}
    \caption{{\textbf{Disease diagnosis performance of LVLMs across different $n$-shot settings}. Each bar corresponds to an $n$-shot scenario ($n = \{0,1,2,4\}$). "0 w/o" corresponds to the 0-shot experiment, where no concepts are used for the disease diagnosis. Filled colored bars denote BACC whereas the hatched bars indicate F1-scores. Red crosses indicate the percentage of unknowns, i.e. the proportion of samples whose LVLMs responses do not contain sufficient information to answer the posed question.}}
    \label{fig:results_disease_diagnosis_performance}
\end{figure*}

This section presents the key results of \emph{CBVLM}, comparing its performance against CBMs and task-specific supervised methods in terms of BACC and F1-score across two tasks: i) Concept Detection ({Figures \ref{fig:results_concept_detection}, \ref{fig:avg_results_concepts}, \ref{fig:spider_plot_concepts_per_dataset}, and \ref{fig:data_percentage}}); and ii) Disease Diagnosis ({Figures \ref{fig:results_disease_diagnosis_performance} and \ref{fig:avg_results_classification}}).

\subsection{Concept Detection Performance}
\label{subsec:concept_detection_performance}
We start by assessing the predictive performance of \emph{CBVLM} in the task of concept detection. Figure~\ref{fig:results_concept_detection} presents the results across different datasets under various $n$-shot settings, $n \in \{0,1,2,4\}$. For comparative analysis, we also report the performance of CBM and CLAT.
The results for $n > 0$ were obtained considering the best vision encoder, in this case, MedImageInsight, selected after an ablation study with different vision encoders (see {Figure \ref{fig:ablation_vision_encoder}}).

\begin{figure*}[!htb]
    \centering
    \begin{tikzpicture}
        \node at (-4.5, 0) {\includegraphics[width=0.395\textwidth]{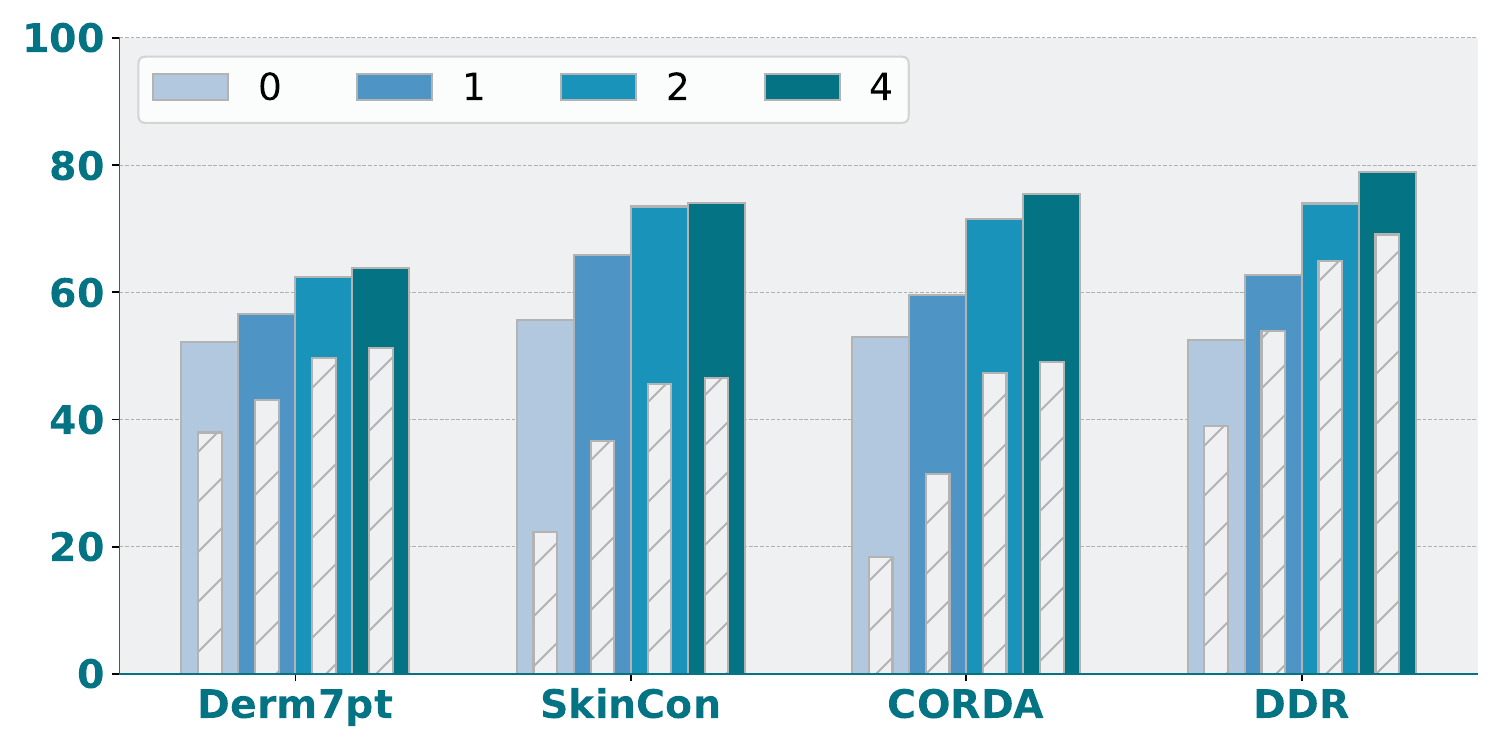}};
        
        \node at (0.25\textwidth, 0.05) {\includegraphics[width=0.56\textwidth]{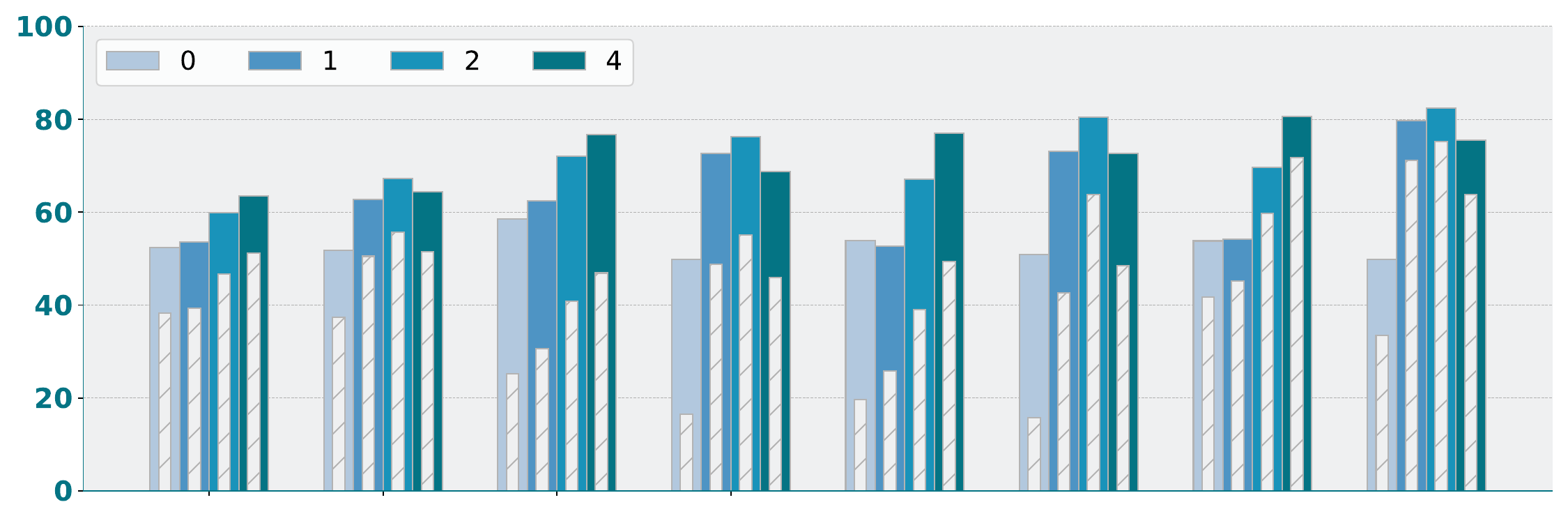}};

        \node[font=\tiny\sffamily\bfseries, text=gray] at (0.75, -1.55) {Generic};
        \node[font=\tiny\sffamily\bfseries, text=gray] at (1.85, -1.55) {Medical};
        \node[font=\scriptsize\sffamily\bfseries, text=promptgreen] at (1.3, -1.85) {Derm7pt};

        \node[font=\tiny\sffamily\bfseries, text=gray] at (2.95, -1.55) {Generic};
        \node[font=\tiny\sffamily\bfseries, text=gray] at (4.05, -1.55) {Medical};
        \node[font=\scriptsize\sffamily\bfseries, text=promptgreen] at (3.6, -1.85) {SkinCon};

        \node[font=\tiny\sffamily\bfseries, text=gray] at (5.1, -1.55) {Generic};
        \node[font=\tiny\sffamily\bfseries, text=gray] at (6.25, -1.55) {Medical};
        \node[font=\scriptsize\sffamily\bfseries, text=promptgreen] at (5.7, -1.85) {CORDA};

        \node[font=\tiny\sffamily\bfseries, text=gray] at (7.3, -1.55) {Generic};
        \node[font=\tiny\sffamily\bfseries, text=gray] at (8.4, -1.55) {Medical};
        \node[font=\scriptsize\sffamily\bfseries, text=promptgreen] at (7.8, -1.85) {DDR};

        \draw[dashed, thick, gray] (2.4, -1.9) -- (2.4, 1.1);
        \draw[dashed, thick, gray] (4.57, -1.9) -- (4.57, 1.1);
        \draw[dashed, thick, gray] (6.73, -1.9) -- (6.73, 1.1);
        
    \end{tikzpicture}
    \caption{{\textbf{Concept detection results per dataset averaged over all models (left) and over generic and medical LVLMs (right).} Each bar corresponds to the number of shots ($n = \{0,1,2,4\}$). Filled colored bars denote BACC, whereas the hatched bars indicate F1-scores.}}
    \label{fig:avg_results_concepts}
\end{figure*}

\paragraph{\textbf{Few-shot prompting boosts performance}} The results presented in Figure~\ref{fig:results_concept_detection} and the analysis depicted in Figure \ref{fig:avg_results_concepts} (left) reveals that few-shot prompting significantly boosts performance by incorporating contextually relevant information within the prompts~\cite{moor2023med}. Notably, there is a consistent trend across all datasets, demonstrating improved performance as the number of demonstrations increases. For example, the performance of \emph{CBVLM} when adopting CheXagent, LLaVA-OV, Qwen2-VL, and SkinGPT-4 improves by over 15\% when incorporating 4 demonstration examples into the prompt compared to the zero-shot scenario ($n = 0$). Remarkably, \emph{CBVLM} outperforms both CBM and CLAT across all datasets except Derm7pt, where its performance is on par with CBM. Impressively, \emph{CBVLM} achieves these results without any training, relying on only four demonstration examples or less.

\paragraph{\textbf{Medical LVLMs outperform generic LVLMs in few-shot settings}} 
From the results depicted in Figure \ref{fig:avg_results_concepts} (right), in the $0$-shot scenario, generic LVLMs consistently outperform medical LVLMs across all datasets. However, in the few-shot setting, medical LVLMs outperform their generic counterparts in all datasets using only one or two demonstration examples ($n \in \{1, 2\}$). The stronger performance of generic LVLMs in the zero-shot setting can be attributed to their training on large-scale generic image datasets, allowing for better generalization in this setting. As demonstration examples are introduced into the prompt, medical LVLMs improve, benefiting from their pre-training on specialized medical data, which enhances their ability to generalize in these scenarios. {However, in the 4-shot setting, generic LVLMs tend to achiever higher concept detection performance than their medical counterparts. This might stem from the fact that generic LVLMs tend to allow for longer prompts and are trained with interleaved image-text data, while medical LVLMs are not.}

\begin{figure*}[!ht]
    \centering
    \includegraphics[width=\textwidth]{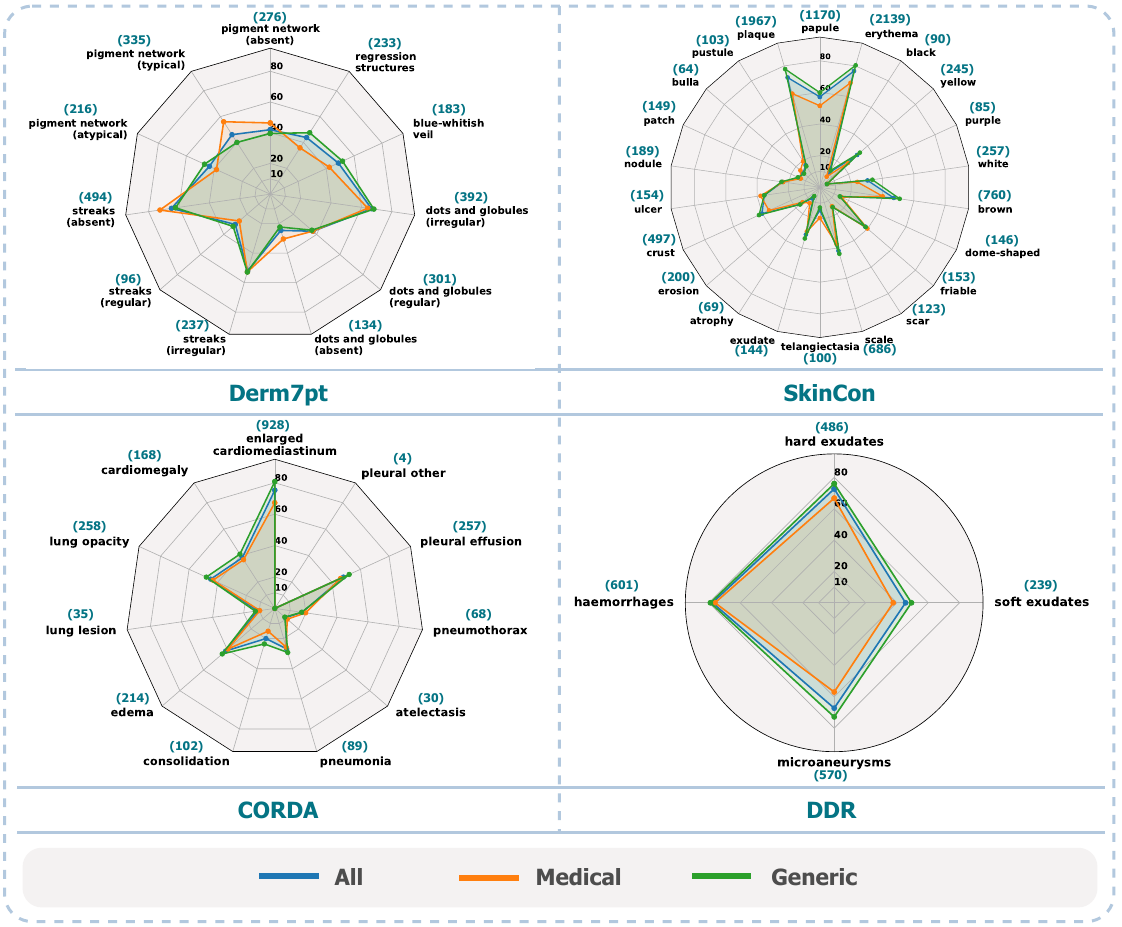} 
    \caption{{\textbf{Per-concept F1-scores across datasets averaged over all (blue), generic (green), and medical (orange) LVLMs, in the 4-shot scenario.} Values in parentheses indicate the number of samples in the dataset for the corresponding concept.}}
    \label{fig:spider_plot_concepts_per_dataset}
\end{figure*}

\begin{figure*}[t]
    \centering
    \begin{tikzpicture}
        \node at (0, 0) {\includegraphics[width=0.25\textwidth]{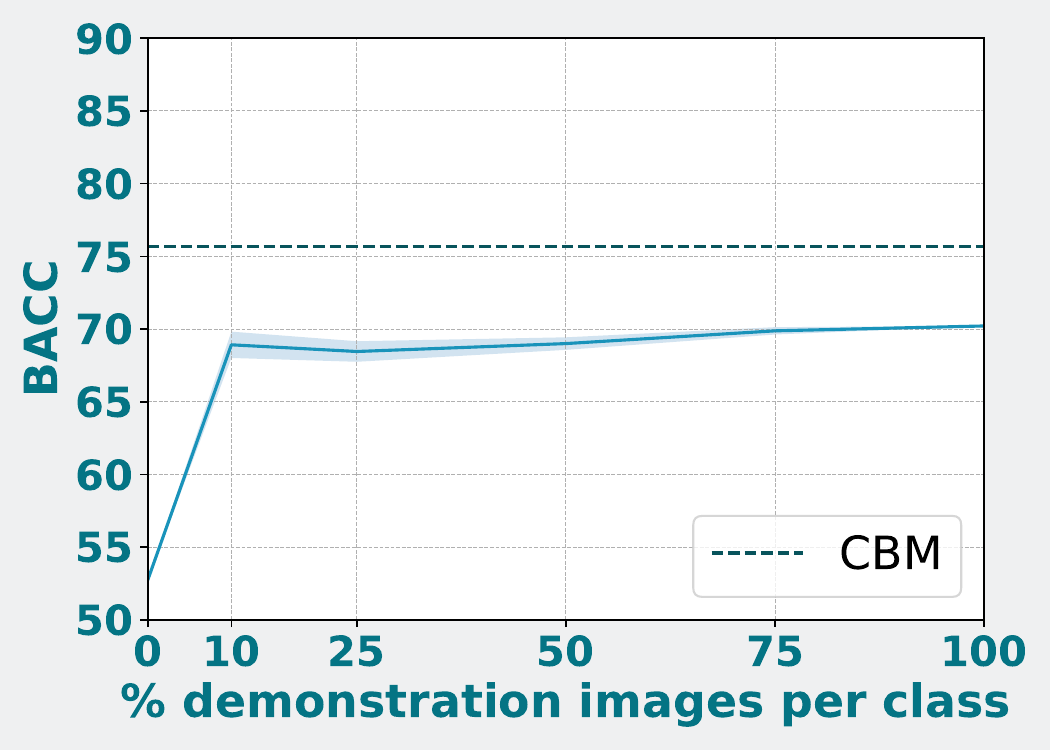}}; 
        \node[font=\footnotesize\sffamily\bfseries, text=black] at (0, -1.8) {Derm7pt};
          
        \node at (0.25\textwidth, 0) {\includegraphics[width=0.25\textwidth]{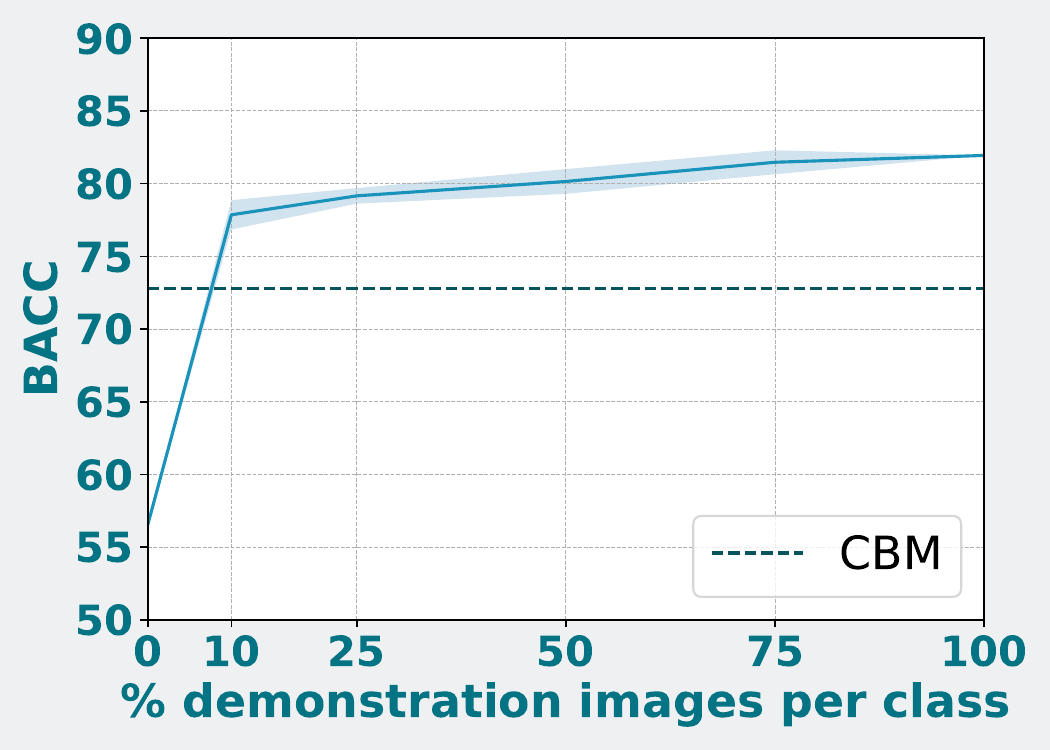 }}; 
        \node[font=\footnotesize\sffamily\bfseries, text=black] at (0.25\textwidth, -1.8) {SkinCon};

        \node at (0.50\textwidth, 0) {\includegraphics[width=0.25\textwidth]{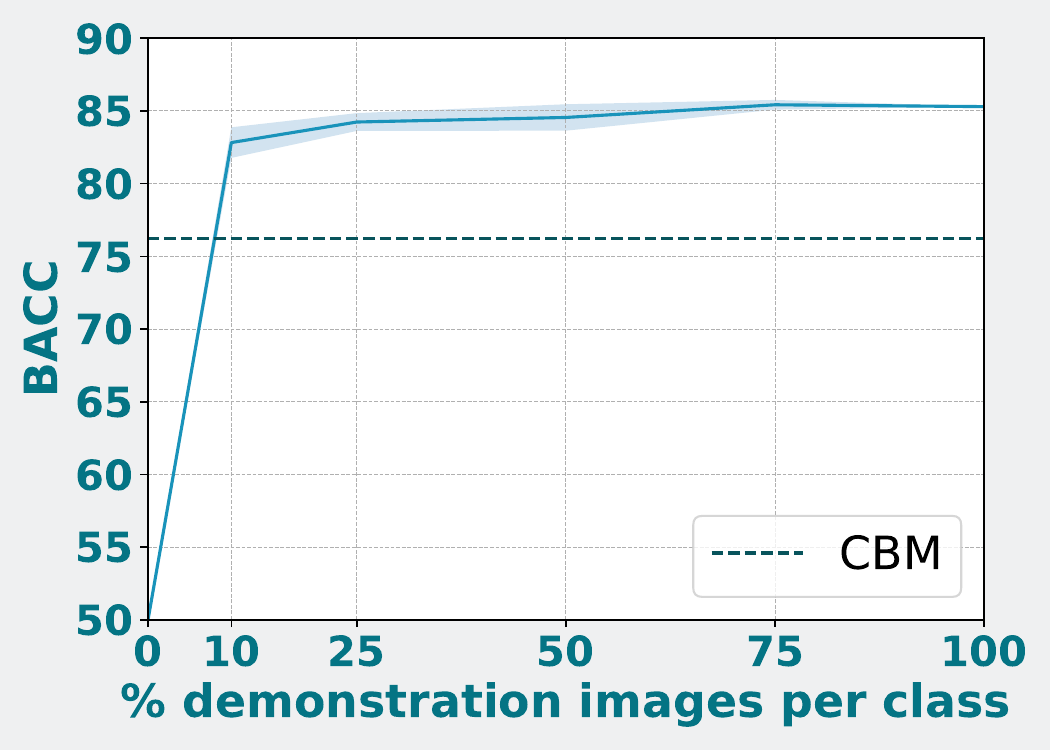}}; 
        \node[font=\footnotesize\sffamily\bfseries, text=black] at (0.5\textwidth, -1.8) {CORDA};

        \node at (0.75\textwidth, 0) {\includegraphics[width=0.25\textwidth]{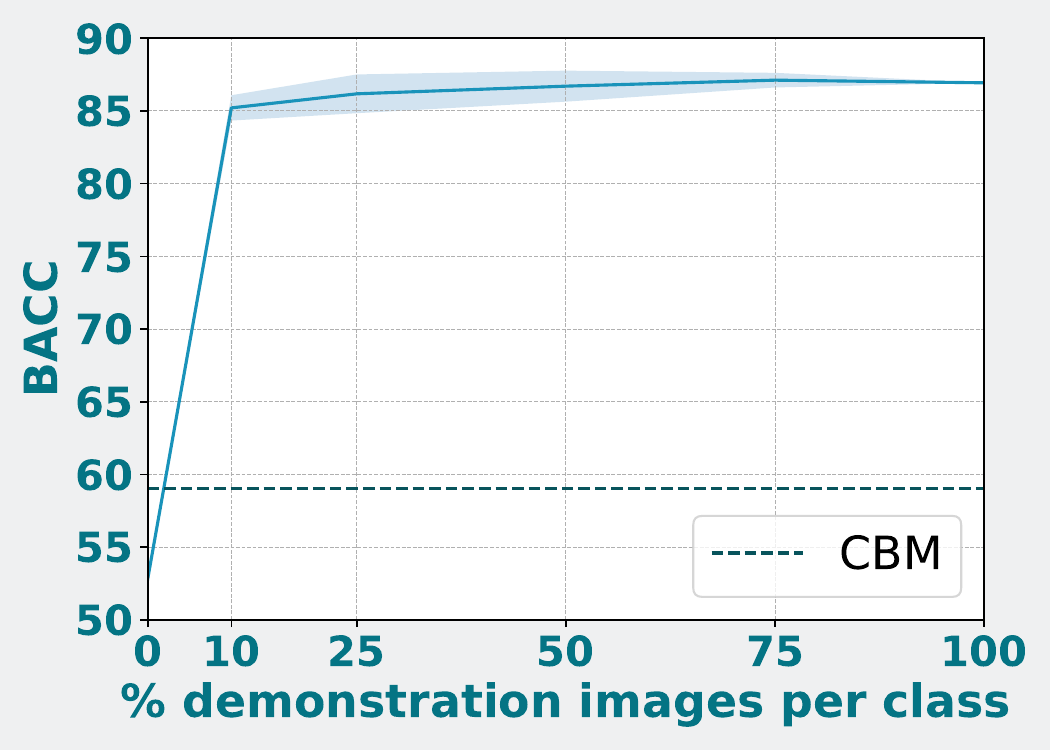}};
        \node[font=\footnotesize\sffamily\bfseries, text=black] at (0.75\textwidth, -1.8) {DDR};
    \end{tikzpicture}
    \vspace{-3mm}
    \caption{{\textbf{Influence of the example set size on the concept prediction performance of \emph{CBVLM}}. Even when only $10\%$ of the example images per class are available, \emph{CBVLM} outperforms CBM, except for Derm7pt. Nevertheless, with only $10\%$ of the examples, the decrease in performance is negligible for Derm7pt. Thus, \emph{CBVLM} indeed requires few annotated examples.}}
    \label{fig:data_percentage}
\end{figure*}

\paragraph{\textbf{Underrepresented concepts are harder to predict}}
Figure~\ref{fig:spider_plot_concepts_per_dataset} presents a detailed comparison of the per-concept F1-score for each dataset. The reported results, averaged across all, generic, and medical LVLMs in the 4-shot setting, reveal that underrepresented concepts are the most challenging to detect, as expected. For instance, in the CORDA dataset, none of the models correctly predicts the presence of the ``pleural other'' concept (which only appears in 4 images of the entire dataset). The concepts ``atelectasis'' and ``lung lesion'' are also consistently difficult to detect. The full results per-LVLM can be found in the Supplementary Material.

\paragraph{\textbf{{CBVLM} outperforms CBMs}} \emph{CBVLM} outperforms CBMs and CLAT in concept prediction across all datasets, except for Derm7pt, where CBM achieves the best result with a margin of $\approx$ 5\%. This may be due to the fact that in Derm7pt, some concepts have three available options, whereas in the other datasets, each concept is binary, which may make it easier for the LVLM to predict the presence or absence of a concept more accurately. 

\paragraph{\textbf{{CBVLM} outperforms CBM when using only 10\% annotated data}}
Although \emph{CBVLM} is able to outperform CBMs with only 4-shots, it should be noted that, in total, more than 4 images are being used, as the same 4 example images will not be the most similar for all query images in the test set.  
To assess the impact of the number of images per class in the examples set, we progressively reduce the percentage of images per class, limiting the number of available examples for selection for few-shot prompting. This experiment is conducted using the best-performing model for each dataset: CheXagent 4-shots for Derm7pt, and Open-Flamingo 2-shots for SkinCon, CORDA and DDR. The results, shown in {Figure \ref{fig:data_percentage}} and averaged across 3 runs, demonstrate that even with just 10\% of the images per class in the examples set, \emph{CBVLM} consistently outperforms CBM across all datasets, except for Derm7pt where even with 100\% of the images per class, the performance is lower than CBM, as discussed earlier. Nevertheless, the results shown in {Figure \ref{fig:data_percentage}} underscore a key observation: few-shot prompting with a good example selection strategy can boost performance by up to 20\% (\textit{cf.} BACC at 0\% and 10\%), even in situations where annotated data is severely limited, which is the case of medical data. Thus, LVLMs can indeed be successfully used for concept detection with very limited amounts of annotated data, thus reducing the annotation cost inherent to CBMs.

\subsection{Disease Diagnosis Performance}

{Figure \ref{fig:results_disease_diagnosis_performance}} shows the performance of \emph{CBVLM} in terms of BACC and F1-score for the task of disease classification across different datasets under various $n$-shots ($n = \{0,1,2,4\}$). Similarly to Section~\ref{subsec:concept_detection_performance}, we also report the performance of CBM, CLAT and a Black-box (Bbox) model for comparison. Additionally, we include results for the setting where the \emph{CBVLM} predicts the disease label directly without prior concept information (``0 w/o'' bar in {Figure \ref{fig:results_disease_diagnosis_performance}}).

\begin{figure*}[!htb]
    \centering
    \begin{tikzpicture}
        \node at (-4.5, 0) {\includegraphics[width=0.395\textwidth]{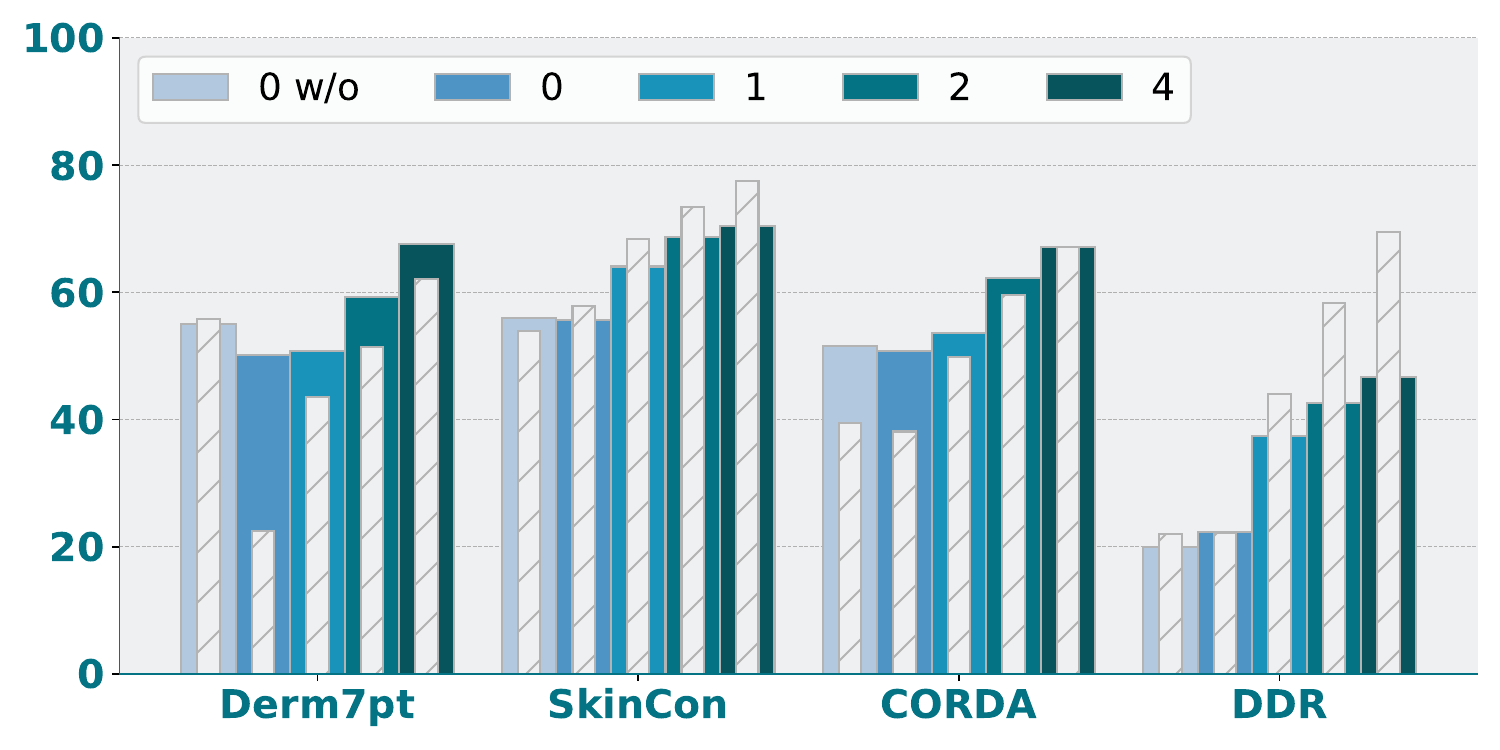}};
        
        \node at (0.25\textwidth, 0.05) {\includegraphics[width=0.56\textwidth]{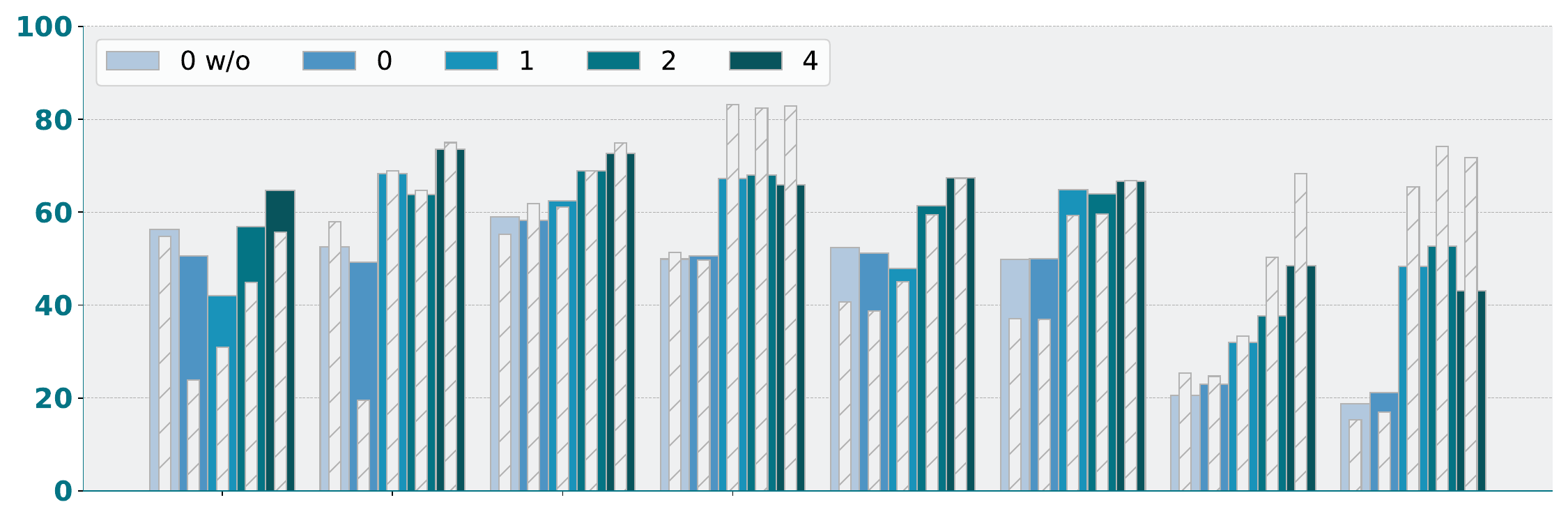}};

        \node[font=\tiny\sffamily\bfseries, text=gray] at (0.8, -1.55) {Generic};
        \node[font=\tiny\sffamily\bfseries, text=gray] at (1.9, -1.55) {Medical};
        \node[font=\scriptsize\sffamily\bfseries, text=promptgreen] at (1.3, -1.85) {Derm7pt};

        \node[font=\tiny\sffamily\bfseries, text=gray] at (2.95, -1.55) {Generic};
        \node[font=\tiny\sffamily\bfseries, text=gray] at (4.05, -1.55) {Medical};
        \node[font=\scriptsize\sffamily\bfseries, text=promptgreen] at (3.6, -1.85) {SkinCon};

        \node[font=\tiny\sffamily\bfseries, text=gray] at (5.1, -1.55) {Generic};
        \node[font=\tiny\sffamily\bfseries, text=gray] at (6.15, -1.55) {Medical};
        \node[font=\scriptsize\sffamily\bfseries, text=promptgreen] at (5.7, -1.85) {CORDA};

        \node[font=\tiny\sffamily\bfseries, text=gray] at (7.2, -1.55) {Generic};
        \node[font=\tiny\sffamily\bfseries, text=gray] at (8.3, -1.55) {Medical};
        \node[font=\scriptsize\sffamily\bfseries, text=promptgreen] at (7.8, -1.85) {DDR};

        \draw[dashed, thick, gray] (2.45, -1.9) -- (2.45, 1.1);
        \draw[dashed, thick, gray] (4.57, -1.9) -- (4.57, 1.1);
        \draw[dashed, thick, gray] (6.7, -1.9) -- (6.7, 1.1);
        
    \end{tikzpicture}
    \caption{{\textbf{Disease diagnosis results per dataset averaged over all models (left) and over generic and medical LVLMs (right).} Each bar corresponds to the number of shots ($n = \{0,1,2,4\}$). "0 w/o" corresponds to the 0-shot experiment where no concepts are used for the disease diagnosis. Filled colored bars denote BACC, whereas the hatched bars indicate F1-scores.}}
    \label{fig:avg_results_classification}
\end{figure*}

\paragraph{\textbf{Few-shot prompting boosts performance}} 
Similar to the trend observed in the concept detection task, the results shown in {Figure \ref{fig:avg_results_classification}} (left) confirm our initial assumption and align with findings reported in the literature: few-shot prompting significantly enhances the performance of LVLMs in disease classification as the number of demonstrations increases, across all datasets.

\paragraph{\textbf{Using concepts improves performance and promotes explainability}} 
The results show that incorporating clinical concepts into the prompt improves both BACC and F1-score compared to the scenario without concept guidance (i.e., ``0'' vs ``0 w/o'' bars in {Figure \ref{fig:results_disease_diagnosis_performance}}). Furthermore, grounding the final diagnosis on a set of clinical concepts promotes transparency and explainability, aligning more closely with human reasoning, unlike the concept-free setting (``0 w/o'').

\paragraph{\textbf{Medical LVLMs outperform generic LVLMs in few-shot
settings}} 
The results in {Figure \ref{fig:avg_results_classification}} (right) confirm that generic LVLMs consistently outperform their medical counterparts across all datasets in the scenario without concepts (``0 w/o''). This trend is also observed in the 0-shot scenario with concepts, in general. 
In the few-shot setting, generic LVLMs exhibit superior performance only in SkinCon in terms of BACC, with a marginal average improvement of 1\% across scenarios with $n > 0$. Conversely, when averaging across all few-shot scenarios ($n > 0$), medical LVLMs consistently outperform generic models across all datasets, achieving F1-score improvements of 25.74\%, 14.48\%, {4.62\%}, and 19.81\% for Derm7pt, SkinCon, CORDA, and DDR, respectively. This phenomenon can be attributed to two factors: i) the stronger domain knowledge of medical LVLMs, and ii) the initialization of their vision encoders from medical-specific CLIP versions, such as BiomedCLIP~\cite{zhang2023biomedclip} for LLaVA-Med, or larger improved CLIP architectures like EVA-CLIP-g~\cite{sun2023eva}, used in CheXagent and SkinGPT.

\paragraph{\textbf{CBVLM outperforms CBMs}}
\emph{CBVLM} surpasses CBMs across all data\-sets. With the top-performing LVLM, we observe F1-score improvements of {7.42\%, 39.70\%, 7.25\%,} and 63.13\% for Derm7pt, SkinCon, CORDA, and DDR, respectively, compared to the standard CBM. It is worth recalling that \emph{CBVLM} achieves these results without any kind of training, whereas CBM needs to be trained for each dataset. 
When comparing \emph{CBVLM} with CLAT~\cite{CLAT}, a method specifically optimized for retinal disease diagnosis, our experiments show that \emph{CBVLM} outperforms it on the diabetic retinopathy dataset, DDR, in terms of F1-score. Furthermore, \emph{CBVLM} also outperforms CLAT on the other datasets, achieving an average relative improvement of $\approx$ 6\%.

\paragraph{\textbf{CBVLM outperforms supervised black-box models}}
Finally, when compared to black-box approaches, a similar trend is observed. \emph{CBVLM} outperforms task-specific black-box models by 2.15\% in BACC for Derm7pt and achieves comparable performance on CORDA, with a slight decrease of {-2.32\%}. In terms of F1 score, the best performing \emph{CBVLM} exceeds black-box task-specific models by 0.36\% on SkinCon and by 1\% on DDR. Thus, \emph{CBVLM} achieves state-of-the-art results on Derm7pt, SkinCon and DDR.

\subsection{Ablation Study: Strategies for ICL Example Selection}
\label{subsec:ablation_vision_encoder}

In order to validate the importance of the \textit{Retrieval Module} (\textit{cf.} Figure \ref{fig:pipeline}), which employs  RICES~\cite{yang2022empirical} (i.e. ranks examples based on the cosine similarity between image features extracted from an arbitrary vision encoder), we conduct an ablation study on the task of concept detection for the $1$-shot scenario. 

\begin{figure*}[!h]
    \centering
    \begin{tabular}{c}
        \vspace{-3mm}
        \begin{tikzpicture}
            \node at (0.0,0.0)
            {\includegraphics[width=0.9\textwidth]{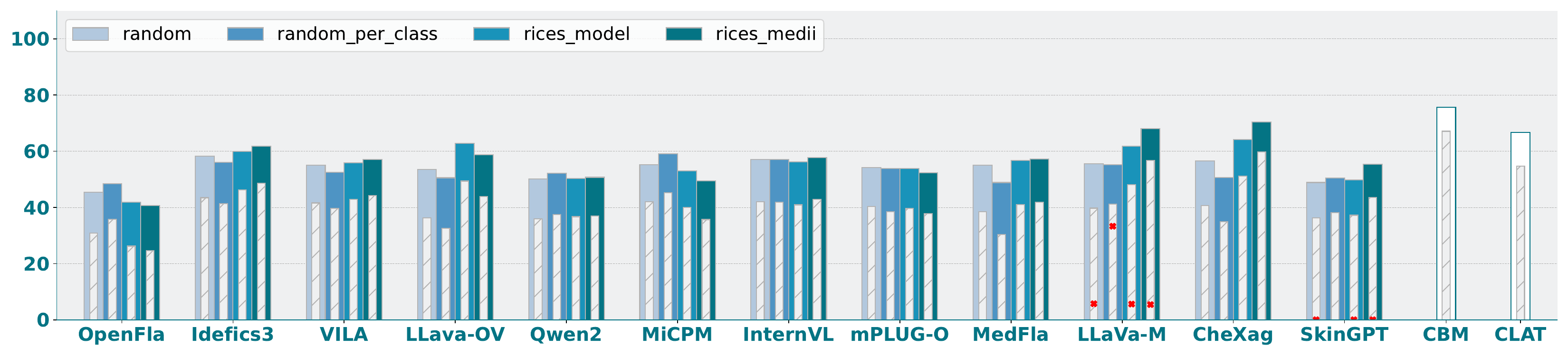}}; 
            \node[font=\footnotesize\sffamily\bfseries, rotate=90, text=black] at (-0.46\textwidth, 0) {Derm7pt};
            \node[font=\footnotesize\sffamily\bfseries, text=gray] at (-0.165\textwidth, 2) {Generic};
            \node[font=\footnotesize\sffamily\bfseries, text=gray] at (0.23\textwidth, 2) {Medical};
            \draw[dashed, thick, gray] (0.099\textwidth, -1.35) -- (0.099\textwidth, 1.645);
            \draw[dashed, thick, gray] (0.355\textwidth, -1.35) -- (0.355\textwidth, 1.645);
        \end{tikzpicture} \\
        \vspace{-3mm}
        \begin{tikzpicture}
            \node at (0, 0) 
            {\includegraphics[width=0.9\textwidth]{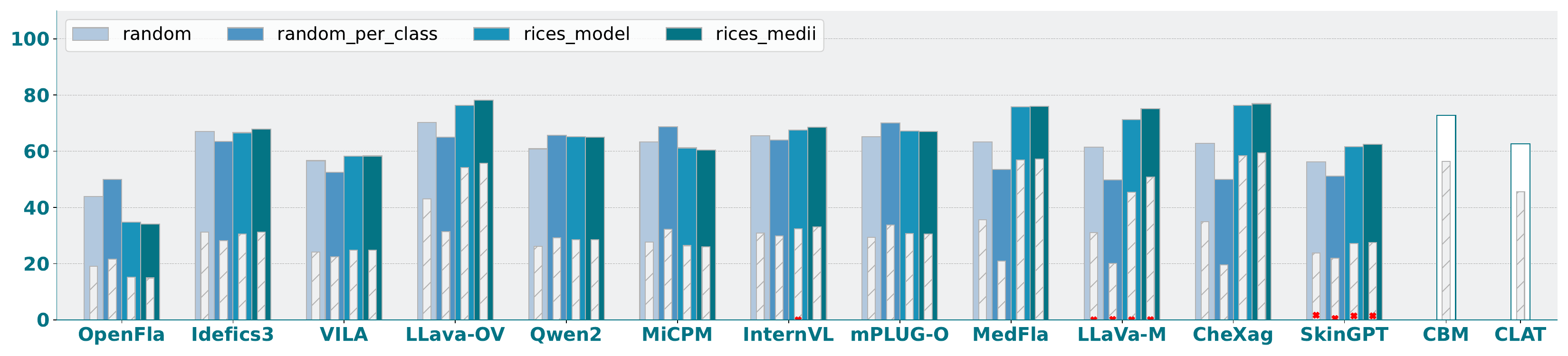}};
            \node[font=\footnotesize\sffamily\bfseries, rotate=90, text=black] at (-0.46\textwidth, 0) {SkinCon};
            \draw[dashed, thick, gray] (0.099\textwidth, -1.35) -- (0.099\textwidth, 1.645);
            \draw[dashed, thick, gray] (0.355\textwidth, -1.35) -- (0.355\textwidth, 1.645);
        \end{tikzpicture} \\
        \vspace{-3mm}
        \begin{tikzpicture}
            \node at (0, 0) 
            {\includegraphics[width=0.9\textwidth]{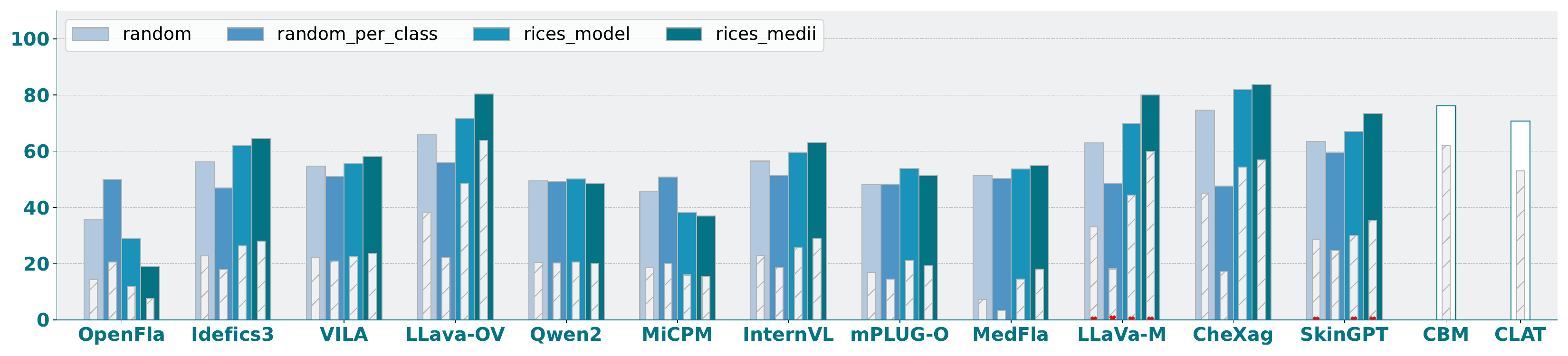}};
            \node[font=\footnotesize\sffamily\bfseries, rotate=90, text=black] at (-0.46\textwidth, 0) {CORDA};
            \draw[dashed, thick, gray] (0.099\textwidth, -1.35) -- (0.099\textwidth, 1.645);
            \draw[dashed, thick, gray] (0.355\textwidth, -1.35) -- (0.355\textwidth, 1.645);
        \end{tikzpicture} \\
        \begin{tikzpicture}
            \node at (0, 0) 
            {\includegraphics[width=0.9\textwidth]{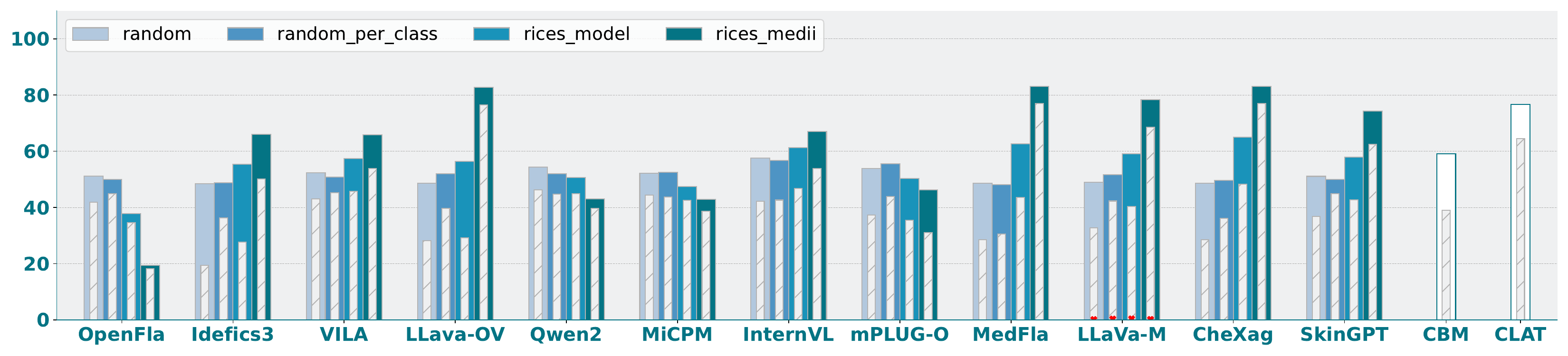}};
            \node[font=\footnotesize\sffamily\bfseries, rotate=90, text=black] at (-0.46\textwidth, 0) {DDR};
            \draw[dashed, thick, gray] (0.099\textwidth, -1.35) -- (0.099\textwidth, 1.645);
            \draw[dashed, thick, gray] (0.355\textwidth, -1.35) -- (0.355\textwidth, 1.645);
        \end{tikzpicture}
    \end{tabular}
    \caption{{\textbf{Concept detection results with 1-shot and several example selection strategies.} ``random'' corresponds to random selection of 1 example, while ``random\_per\_class'' corresponds to selecting 1 example per class randomly. ``rices'' is related to the RICES methodology~\cite{yang2022empirical}, where the demonstration example is chosen based on the cosine similarity to the query image's features. ``rices\_medii'' corresponds to using RICES when MedImageInsight~\cite{codella2024medimageinsight} is the vision encoder used to extract the features from the query and demonstration images, while in ``rices\_model'' the encoder is the vision model included in the corresponding LVLM.}}
    \label{fig:ablation_vision_encoder}
\end{figure*}

In this study, we first compare random ICL example selection with RICES, and then assess different vision encoders to perform RICES. For random selection, we compare choosing only 1 random example against choosing 1 random example per class. The results of {Figure \ref{fig:ablation_vision_encoder}} show that, except for Open-Flamingo and MiniCPM-V, using 1 example per class does not improve performance. In fact, in the majority of cases it slightly decreases it, when comparing to simply choose 1 random example.

We then explore two different encoders for RICES: MedImageInsight~\cite{codella2024medimageinsight} and the encoder integrated within the respective LVLM. Although in some models (e.g., Open-Flamingo and MiniCPM-V), choosing the examples randomly brings better performance, in the majority of cases it improves both BACC and F1-score (especially for Medical LVLMs). More importantly, we are not concerned with the average results, but rather on maximizing them, even if just for one LVLM, to be able to find the best configuration of \emph{CBVLM} to compete against other approaches like CBM and supervised black-box models. Thus, using RICES beats random selection and using MedImageInsight instead of the LVLM encoder yields the best results on all datasets. As such, all experiments are conducted using RICES (i.e. the \textit{Retrieval Module}) with MedImageInsight as the feature extractor. 

\subsection{General Discussion}
Table \ref{tab:comparison_benchmark} presents a detailed comparison of our proposed \emph{CBVLM} with concept-based models, supervised black-box models, and task-specific models. As highlighted in the results, \emph{CBVLM} consistently achieves either the best or second-best performance across all evaluations. Notably, this is accomplished without any additional training and by leveraging just four examples, enabling concept-based explanations and enhanced interpretability and transparency in the model responses. In summary, \emph{CBVLM} using CheXagent with 4 demonstration examples emerges as the top-performing combination for Derm7pt. For SkinCon, \emph{CBVLM} with the generic Open-Flamingo model and 4 demonstration examples achieves the best results in both disease diagnosis and concept detection tasks. {For CORDA, the best performance is achieved by \emph{CBVLM} with CheXagent with 2 demonstration examples.} For DDR, \emph{CBVLM} with Idefics3 and 4 demonstrations proves to be the best-performing combination.

Overall, the results underscore the effectiveness of \emph{CBVLM} in enabling arbitrary LVLMs to predict the presence of clinical concepts from an image and use those concepts to ground the final diagnosis, enhancing interpretability and transparency in the model's responses, all without requiring any training and with just a few annotated examples. Notably, \emph{CBVLM} outperforms CBM-related approaches across all datasets and even surpasses black-box models when employing ICL.

\begin{table}
\caption{\textbf{Comparison of the best performing \emph{CBVLM} with state-of-the-art supervised models and CBMs for disease diagnosis and concept detection}. BACC refers to Balanced Accuracy, and F1 refers to the F1-score. The best performance is highlighted in \textbf{bold}, while the second-best performance is indicated with \underline{underlining}.}
\label{tab:comparison_benchmark}
\setlength{\tabcolsep}{3.5pt}
\begin{center}
\begin{adjustbox}{width=0.82\columnwidth}
\begin{tabular}{lcccccc}
\toprule
\multirow{2}{*}{\textbf{Dataset}} & \multirow{2}{*}{\textbf{Method}} & \multicolumn{2}{c}{\textbf{Disease Diagnosis}} & \multicolumn{2}{c}{\textbf{Concept Detection}} \\
& & \textbf{\scriptsize{BACC (\%)}} & \textbf{\scriptsize{F1 (\%)}} & \textbf{\scriptsize{BACC (\%)}} & \textbf{\scriptsize{F1 (\%)}} \\
\midrule
\multirow{6}{*}{Derm7pt} & CBM~\cite{koh2020concept} & 74.01 & 75.66 & \textbf{75.68}& \textbf{67.11} \\
& CLAT~\cite{CLAT} & 79.67 & \underline{82.98} & 66.69 & 54.76 \\ 
& Black-box (ResNet50) & 75.33 & 80.76 & - & - \\
& Black-box (ViT Base) & 82.26 & 82.04 & - & - \\
& Black-box (Task-Specific)~\cite{patricio2024towards} & \underline{83.20} & - & - & - \\
& \cellcolor{mycolor} \textbf{\emph{CBVLM}} {\scriptsize{(CheXagent \& $4$-shot)}} & \cellcolor{mycolor} \textbf{85.35} & \cellcolor{mycolor} \textbf{83.08} & \cellcolor{mycolor} \underline{70.21} & \cellcolor{mycolor} \underline{59.55} \\
\midrule
\multirow{6}{*}{SkinCon} & CBM~\cite{koh2020concept} & 49.97 & 46.51 & \underline{72.81} & \underline{56.33} \\
& CLAT~\cite{CLAT} & 68.21 & \textbf{86.76} & 62.64 & 45.52 \\
& Black-box (ResNet50) & 73.98 & 81.49 & - & - \\
& Black-box (ViT Base) & \underline{80.25} & 85.85 & - & - \\
& Black-box (Task-Specific)~\cite{hou2024concept} & - & 77.80 & - & - \\
& \cellcolor{mycolor} \textbf{\emph{CBVLM}} {\scriptsize{(Open-Flamingo \& $4$-shot)}} & \cellcolor{mycolor} \textbf{85.61} & \cellcolor{mycolor} \underline{86.21} & \cellcolor{mycolor} \textbf{79.78} & \cellcolor{mycolor} \textbf{57.91} \\
\midrule
\multirow{6}{*}{CORDA} 
& {CBM ~\cite{koh2020concept}} & {70.75} & {70.98} & {\underline{76.20}} & {\underline{61.92}} \\
& {CLAT ~\cite{CLAT}} & {75.64} & {\underline{77.96}} & {70.76} & {53.06}\\
& Black-box (ResNet50) & 73.17 & 73.28 & - & - \\
& Black-box (ViT Base) & \underline{78.03} & {77.63} & - & - \\
& Black-box (Task-Specific)~\cite{barbano2024ai} & \textbf{80.00} & - & - & - \\
& \cellcolor{mycolor} {\textbf{\emph{CBVLM}} {\scriptsize{(CheXagent \& $2$-shot)}}} & \cellcolor{mycolor} {77.68} & \cellcolor{mycolor} {\textbf{78.23}} & \cellcolor{mycolor} {\textbf{81.38}} & \cellcolor{mycolor} {\textbf{64.33}}\\
\midrule
\multirow{6}{*}{DDR} & CBM~\cite{koh2020concept} & 23.62 & 21.25 & 59.05 & 39.06 \\
& CLAT~\cite{CLAT} & \textbf{78.87} & 72.81 & \underline{76.64} & \underline{64.53}\\
& Black-box (ResNet50) & 33.70 & 57.00 & - & - \\
& Black-box (ViT Base) & 39.09 & 59.37 & - & - \\
& Black-box (Task-Specific)~\cite{madarapu2024multi} & - & \underline{83.38} & - & - \\
& \cellcolor{mycolor} \textbf{\emph{CBVLM}} {\scriptsize{(Idefics3 \& $4$-shot)}} & \cellcolor{mycolor} \underline{66.65} & \cellcolor{mycolor} \textbf{84.38} & \cellcolor{mycolor} \textbf{82.69} & \cellcolor{mycolor} \textbf{75.73} \\
\bottomrule
\end{tabular}
\end{adjustbox}
\end{center}
\end{table}
\subsection{Limitations}
The main drawback of the proposed methodology for generating concept-based explanations with LVLMs lies in the independent prediction of each concept, which is a concern when applied to datasets with a large number of concepts. A similar problem arises with datasets with a large number of classes, as each class is provided as a separate option in the prompt. This can result in excessively long prompts that would require alternative ways of querying the model. {Although in medical applications the number of concepts is typically limited, a potential future direction to address concerns regarding the scalability for datasets with numerous concepts could be the adoption of tailored selection policies that prioritize only a small subset of the most informative concepts, as demonstrated in recent works~\cite{chauhan2023interactive}.}

Additionally, our approach is limited to discrete-valued concepts (e.g., absent/present or, for some concepts of Derm7pt, absent/regular/irregular), contrasting with traditional CBMs, where each concept is associated with a probability. Nevertheless, this limitation could be addressed by prompting the LVLM to not only answer if a given concept is present in the input image but also to provide a confidence score, as is done in other works~\cite{ferber2024context}. We leave this exploration to future work.

Finally, \emph{CBVLM} is only as good as the LVLM it uses. In a few cases, some of the LVLMs we tested struggled to provide answers to the questions (e.g., instead of answering the question, the LVLM simply described the input image), leading to an increase in unknown responses (as shown in Figures \ref{fig:results_concept_detection} and \ref{fig:results_disease_diagnosis_performance}). This phenomenon is particularly common with SkinGPT, which may be due to its pretraining on very specific prompts designed to elicit desired responses~\cite{zhou2024pre}. LLava-Med also faced challenges in providing valid responses, especially in few-shot scenarios where $n > 1$. This difficulty may stem from the fact that it is not trained with interleaved image-text data; in such models, image data is given at the input in the form of tokens, so as the number of shots increases, so does the number of tokens, making it more difficult for the model to pay attention to the full input sequence~\cite{li2024long}.

{Another relevant concern involves the robustness, security, and inherent biases of LVLMs~\cite{zhang2024b}. While a comprehensive analysis of these aspects is beyond the scope of this work, it remains a critical consideration for future studies aimed at safe and reliable integration of LVLMs into clinical workflows. Collaboration between clinicians and AI researchers is crucial to ensure transparency and explainability of AI models deployed into clinical practice~\cite{phung2023best}.}

\section{Conclusions and Future Work}
We propose a simple yet effective methodology, \emph{CBVLM}, that leverages off-the-shelf LVLMs to produce concept-based explanations and predict disease diagnoses grounded on those explanations, thus offering a more transparent and explainable decision-making process, something that is critical in high-stakes scenarios such as medical use-cases. Through extensive experiments across four medical datasets, and open-source LVLMs, we show that \emph{CBVLM} outperforms the traditional CBM and even task-specific black-box models across all benchmarks. This is achieved without requiring any training and using just a few annotated examples, thus tackling a very common problem in medical applications, where the annotation process requires a clinician's expertise. Overall, \emph{CBVLM} is able to leverage the best of both worlds: (i) its LVLM-based training-free nature presents a major advantage over CBMs, as these models require additional training and are restricted by predefined concepts, and (ii) its two-stage process keeps the interpretability inherent to CBMs but that LVLMs lack.

Although \emph{CBVLM} was tested on medical datasets, it is a general methodology that can be applied in other domains. In the future, we would like to do so, e.g., on the CUB dataset~\cite{cubdataset}. Given the promising results obtained in this work with open-source LVLMs of around 8 billion parameters, we would also like to experiment with bigger and/or proprietary LVLMs.

\bibliographystyle{IEEEtran}
\bibliography{main}

\begin{thebibliography}{10}
\providecommand{\url}[1]{#1}
\csname url@samestyle\endcsname
\providecommand{\newblock}{\relax}
\providecommand{\bibinfo}[2]{#2}
\providecommand{\BIBentrySTDinterwordspacing}{\spaceskip=0pt\relax}
\providecommand{\BIBentryALTinterwordstretchfactor}{4}
\providecommand{\BIBentryALTinterwordspacing}{\spaceskip=\fontdimen2\font plus
\BIBentryALTinterwordstretchfactor\fontdimen3\font minus \fontdimen4\font\relax}
\providecommand{\BIBforeignlanguage}[2]{{%
\expandafter\ifx\csname l@#1\endcsname\relax
\typeout{** WARNING: IEEEtran.bst: No hyphenation pattern has been}%
\typeout{** loaded for the language `#1'. Using the pattern for}%
\typeout{** the default language instead.}%
\else
\language=\csname l@#1\endcsname
\fi
#2}}
\providecommand{\BIBdecl}{\relax}
\BIBdecl

\bibitem{moor2023med}
M.~Moor, Q.~Huang, S.~Wu, M.~Yasunaga, Y.~Dalmia, J.~Leskovec, C.~Zakka, E.~P. Reis, and P.~Rajpurkar, ``{Med-Flamingo: a Multimodal Medical Few-shot Learner},'' in \emph{Proceedings of the 3rd Machine Learning for Health Symposium}, ser. Proceedings of Machine Learning Research, vol. 225.\hskip 1em plus 0.5em minus 0.4em\relax PMLR, 2023, pp. 353--367.

\bibitem{li2024llava}
C.~Li, C.~Wong, S.~Zhang, N.~Usuyama, H.~Liu, J.~Yang, T.~Naumann, H.~Poon, and J.~Gao, ``{LLaVA-Med: Training a Large Language-and-Vision Assistant for Biomedicine in One Day},'' \emph{Advances in Neural Information Processing Systems (NeurIPS)}, vol.~36, 2024.

\bibitem{chen2024chexagent}
Z.~Chen, M.~Varma, J.-B. Delbrouck, M.~Paschali, L.~Blankemeier, D.~Van~Veen, J.~M.~J. Valanarasu, A.~Youssef, J.~P. Cohen, E.~P. Reis \emph{et~al.}, ``{CheXagent: Towards a Foundation Model for Chest X-Ray Interpretation},'' \emph{arXiv:2401.12208}, 2024.

\bibitem{zhou2024pre}
J.~Zhou, X.~He, L.~Sun, J.~Xu, X.~Chen, Y.~Chu, L.~Zhou, X.~Liao, B.~Zhang, S.~Afvari \emph{et~al.}, ``{Pre-trained multimodal large language model enhances dermatological diagnosis using SkinGPT-4},'' \emph{Nature Communications}, vol.~15, no.~1, p. 5649, 2024.

\bibitem{han2023multimodal}
T.~Han, L.~C. Adams, S.~Nebelung, J.~N. Kather, K.~K. Bressem, and D.~Truhn, ``{Multimodal Large Language Models are Generalist Medical Image Interpreters},'' \emph{medRxiv}, pp. 2023--12, 2023.

\bibitem{van2024large}
M.-H. Van, P.~Verma, and X.~Wu, ``{On Large Visual Language Models for Medical Imaging Analysis: An Empirical Study},'' in \emph{2024 IEEE/ACM Conference on Connected Health: Applications, Systems and Engineering Technologies (CHASE)}, 2024, pp. 172--176.

\bibitem{ferber2024context}
D.~Ferber, G.~W{\"o}lflein, I.~C. Wiest, M.~Ligero, S.~Sainath, N.~Ghaffari~Laleh, O.~S. El~Nahhas, G.~M{\"u}ller-Franzes, D.~J{\"a}ger, D.~Truhn \emph{et~al.}, ``{In-context learning enables multimodal large language models to classify cancer pathology images},'' \emph{Nature Communications}, vol.~15, no.~1, p. 10104, 2024.

\bibitem{rudin2019stop}
C.~Rudin, ``{Stop explaining black box machine learning models for high stakes decisions and use interpretable models instead},'' \emph{Nature Machine Intelligence}, vol.~1, no.~5, pp. 206--215, 2019.

\bibitem{liu2023llava}
H.~Liu, C.~Li, Q.~Wu, and Y.~J. Lee, ``{Visual Instruction Tuning},'' in \emph{Advances in Neural Information Processing Systems (NeurIPS)}, vol.~36, 2023, pp. 34\,892--34\,916.

\bibitem{OpenAI_GPT4_2023}
OpenAI, ``{GPT-4 Technical Report},'' \emph{arXiv:2303.08774}, 2023.

\bibitem{royer2024multimedeval}
C.~Royer, B.~Menze, and A.~Sekuboyina, ``{MultiMedEval: A Benchmark and a Toolkit for Evaluating Medical Vision-Language Models},'' \emph{arXiv:2402.09262}, 2024.

\bibitem{wang2024asclepius}
J.~Liu, W.~Wang, S.~Yihang, J.~Huang, Y.~Zhang, C.-Y. Li, W.~Chen, X.~Xing, K.-J. Chang, L.~Shen, and M.~R. Lyu, ``{Asclepius: A Spectrum Evaluation Benchmark for Medical Multi-Modal Large Language Models},'' in \emph{Proceedings of the 63rd Annual Meeting of the Association for Computational Linguistics (Volume 1: Long Papers)}, 2025, pp. 24\,181--24\,201.

\bibitem{xia2024cares}
P.~Xia, Z.~Chen, J.~Tian, Y.~Gong, R.~Hou, Y.~Xu, Z.~Wu, Z.~Fan, Y.~Zhou, K.~Zhu, W.~Zheng, Z.~Wang, X.~Wang, X.~Zhang, C.~Bansal, M.~Niethammer, J.~Huang, H.~Zhu, Y.~Li, J.~Sun, Z.~Ge, G.~Li, J.~Zou, and H.~Yao, ``{CARES: A Comprehensive Benchmark of Trustworthiness in Medical Vision Language Models},'' in \emph{Advances in Neural Information Processing Systems (NeurIPS)}, vol.~37, 2024, pp. 140\,334--140\,365.

\bibitem{koh2020concept}
P.~W. Koh, T.~Nguyen, Y.~S. Tang, S.~Mussmann, E.~Pierson, B.~Kim, and P.~Liang, ``{Concept Bottleneck Models},'' in \emph{Proceedings of the International Conference on Machine Learning (ICML)}.\hskip 1em plus 0.5em minus 0.4em\relax PMLR, 2020, pp. 5338--5348.

\bibitem{patricio2023coherent}
C.~Patr{\'\i}cio, J.~C. Neves, and L.~F. Teixeira, ``{Coherent Concept-Based Explanations in Medical Image and Its Application to Skin Lesion Diagnosis},'' in \emph{Proceedings of the IEEE/CVF Conference on Computer Vision and Pattern Recognition Workshops (CVPRW)}, 2023, pp. 3798--3807.

\bibitem{bie2024mica}
Y.~Bie, L.~Luo, and H.~Chen, ``{MICA: Towards Explainable Skin Lesion Diagnosis via Multi-Level Image-Concept Alignment},'' in \emph{Proceedings of the AAAI Conference on Artificial Intelligence}, vol.~38, no.~2, 2024, pp. 837--845.

\bibitem{patricio2024towards}
C.~Patr{\'\i}cio, L.~F. Teixeira, and J.~C. Neves, ``{Towards Concept-based Interpretability of Skin Lesion Diagnosis using Vision-Language Models},'' in \emph{Proceedings of the IEEE International Symposium on Biomedical Imaging (ISBI)}.\hskip 1em plus 0.5em minus 0.4em\relax IEEE, 2024, pp. 1--5.

\bibitem{ghosh2023bridging}
S.~Ghosh, K.~Yu, F.~Arabshahi, and K.~Batmanghelich, ``{Bridging the Gap: From Post Hoc Explanations to Inherently Interpretable Models for Medical Imaging},'' in \emph{Proceedings of the ICML Workshop on Interpretable Machine Learning in Healthcare}, 2023.

\bibitem{OpenAI_ChatGPT}
OpenAI, ``{ChatGPT: Optimizing Language Models for Dialogue},'' \url{https://chat.openai.com/}, 2024.

\bibitem{gpt3}
T.~Brown, B.~Mann, N.~Ryder, M.~Subbiah, J.~D. Kaplan, P.~Dhariwal, A.~Neelakantan, P.~Shyam, G.~Sastry, A.~Askell, S.~Agarwal, A.~Herbert-Voss, G.~Krueger, T.~Henighan, R.~Child, A.~Ramesh, D.~Ziegler, J.~Wu, C.~Winter, C.~Hesse, M.~Chen, E.~Sigler, M.~Litwin, S.~Gray, B.~Chess, J.~Clark, C.~Berner, S.~McCandlish, A.~Radford, I.~Sutskever, and D.~Amodei, ``{Language Models are Few-Shot Learners},'' in \emph{Advances in Neural Information Processing Systems (NeurIPS)}, vol.~33, 2020, pp. 1877--1901.

\bibitem{alayrac2022flamingo}
J.-B. Alayrac, J.~Donahue, P.~Luc, A.~Miech, I.~Barr, Y.~Hasson, K.~Lenc, A.~Mensch, K.~Millican, M.~Reynolds \emph{et~al.}, ``{Flamingo: a Visual Language Model for Few-Shot Learning},'' in \emph{Advances in Neural Information Processing Systems (NeurIPS)}, vol.~35, 2022, pp. 23\,716--23\,736.

\bibitem{zhou2024adapting}
G.~Zhou, Z.~Han, S.~Chen, B.~Huang, L.~Zhu, S.~Khan, X.~Gao, and L.~Yao, ``{Adapting Large Multimodal Models to Distribution Shifts: The Role of In-Context Learning},'' \emph{arXiv:2405.12217}, 2024.

\bibitem{awadalla2023openflamingo}
A.~Awadalla, I.~Gao, J.~Gardner, J.~Hessel, Y.~Hanafy, W.~Zhu, K.~Marathe, Y.~Bitton, S.~Gadre, S.~Sagawa \emph{et~al.}, ``{OpenFlamingo: An Open-Source Framework for Training Large Autoregressive Vision-Language Models},'' \emph{arXiv:2308.01390}, 2023.

\bibitem{yang2022empirical}
Z.~Yang, Z.~Gan, J.~Wang, X.~Hu, Y.~Lu, Z.~Liu, and L.~Wang, ``{An Empirical Study of GPT-3 for Few-Shot Knowledge-Based VQA},'' in \emph{Proceedings of the AAAI Conference on Artificial Intelligence}, vol.~36, no.~3, 2022, pp. 3081--3089.

\bibitem{chen2024understanding}
S.~Chen, Z.~Han, B.~He, M.~Buckley, P.~Torr, V.~Tresp, and J.~Gu, ``{Understanding and Improving In-Context Learning on Vision-language Models},'' in \emph{ICLR 2024 Workshop on Mathematical and Empirical Understanding of Foundation Models}, 2024.

\bibitem{wang2024-answerc}
X.~Wang, B.~Ma, C.~Hu, L.~Weber-Genzel, P.~R{\"o}ttger, F.~Kreuter, D.~Hovy, and B.~Plank, ``{{\textquotedblleft}My Answer is {C}{\textquotedblright}: First-Token Probabilities Do Not Match Text Answers in Instruction-Tuned Language Models},'' in \emph{Findings of the Association for Computational Linguistics: ACL 2024}, 2024, pp. 7407--7416.

\bibitem{DERM7PT}
J.~Kawahara, S.~Daneshvar, G.~Argenziano, and G.~Hamarneh, ``{Seven-Point Checklist and Skin Lesion Classification Using Multitask Multimodal Neural Nets},'' \emph{IEEE Journal of Biomedical and Health Informatics (JBHI)}, vol.~23, no.~2, pp. 538--546, 2019.

\bibitem{daneshjou2022skincon}
R.~Daneshjou, M.~Yuksekgonul, Z.~R. Cai, R.~Novoa, and J.~Y. Zou, ``{SkinCon: A skin disease dataset densely annotated by domain experts for fine-grained model debugging and analysis},'' in \emph{Advances in Neural Information Processing Systems (NeurIPS)}, vol.~35, 2022, pp. 18\,157--18\,167.

\bibitem{cordadataset}
M.~Alesina, C.~A. Barbano, C.~Berzovini, M.~Busso, M.~Calandri, A.~D. Pascale, A.~Fiandrotti, P.~Fonio, M.~Grangetto \emph{et~al.}, ``{CORDA Dataset},'' Apr. 2023.

\bibitem{li2019diagnostic}
T.~Li, Y.~Gao, K.~Wang, S.~Guo, H.~Liu, and H.~Kang, ``Diagnostic assessment of deep learning algorithms for diabetic retinopathy screening,'' \emph{Information Sciences}, vol. 501, pp. 511--522, 2019.

\bibitem{groh2021evaluating}
M.~Groh, C.~Harris, L.~Soenksen, F.~Lau, R.~Han, A.~Kim, A.~Koochek, and O.~Badri, ``{Evaluating Deep Neural Networks Trained on Clinical Images in Dermatology with the Fitzpatrick 17k Dataset},'' in \emph{2021 IEEE/CVF Conference on Computer Vision and Pattern Recognition Workshops (CVPRW)}, 2021, pp. 1820--1828.

\bibitem{barbano2022two}
C.~A. Barbano, E.~Tartaglione, C.~Berzovini, M.~Calandri, and M.~Grangetto, ``{A Two-Step Radiologist-Like Approach for Covid-19 Computer-Aided Diagnosis from Chest X-Ray Images},'' in \emph{Image Analysis and Processing -- ICIAP 2022}, 2022, pp. 173--184.

\bibitem{irvin2019chexpert}
J.~Irvin, P.~Rajpurkar, M.~Ko, Y.~Yu, S.~Ciurea-Ilcus, C.~Chute, H.~Marklund, B.~Haghgoo, R.~Ball, K.~Shpanskaya \emph{et~al.}, ``{CheXpert: A Large Chest Radiograph Dataset with Uncertainty Labels and Expert Comparison},'' in \emph{Proceedings of the AAAI Conference on Artificial Intelligence}, vol.~33, no.~01, 2019, pp. 590--597.

\bibitem{idefics3}
H.~Lauren{\c{c}}on, A.~Marafioti, V.~Sanh, and L.~Tronchon, ``{Building and better understanding vision-language models: insights and future directions},'' in \emph{Workshop on Responsibly Building the Next Generation of Multimodal Foundational Models}, 2024.

\bibitem{lin2024vila}
J.~Lin, H.~Yin, W.~Ping, P.~Molchanov, M.~Shoeybi, and S.~Han, ``{VILA: On Pre-training for Visual Language Models},'' in \emph{Proceedings of the IEEE/CVF Conference on Computer Vision and Pattern Recognition (CVPR)}, 2024, pp. 26\,689--26\,699.

\bibitem{li2024llavaov}
B.~Li, Y.~Zhang, D.~Guo, R.~Zhang, F.~Li, H.~Zhang, K.~Zhang, P.~Zhang, Y.~Li, Z.~Liu, and C.~Li, ``{{LL}a{VA}-OneVision: Easy Visual Task Transfer},'' \emph{Transactions on Machine Learning Research}, 2025.

\bibitem{Qwen-VL}
J.~Bai, S.~Bai, S.~Yang, S.~Wang, S.~Tan, P.~Wang, J.~Lin, C.~Zhou, and J.~Zhou, ``{Qwen-VL: A Versatile Vision-Language Model for Understanding, Localization, Text Reading, and Beyond},'' \emph{arXiv:2308.12966}, 2023.

\bibitem{yao2024minicpm}
Y.~Yao, T.~Yu, A.~Zhang, C.~Wang, J.~Cui, H.~Zhu, T.~Cai, H.~Li, W.~Zhao, Z.~He \emph{et~al.}, ``{MiniCPM-V: A GPT-4V Level MLLM on Your Phone},'' \emph{arXiv:2408.01800}, 2024.

\bibitem{chen2025internvl25}
Z.~Chen, W.~Wang, Y.~Cao, Y.~Liu, Z.~Gao, E.~Cui, J.~Zhu, S.~Ye, H.~Tian, Z.~Liu, L.~Gu, X.~Wang, Q.~Li, Y.~Ren, Z.~Chen, J.~Luo, J.~Wang, T.~Jiang, B.~Wang, C.~He, B.~Shi, X.~Zhang, H.~Lv, Y.~Wang, W.~Shao, P.~Chu, Z.~Tu, T.~He, Z.~Wu, H.~Deng, J.~Ge, K.~Chen, K.~Zhang, L.~Wang, M.~Dou, L.~Lu, X.~Zhu, T.~Lu, D.~Lin, Y.~Qiao, J.~Dai, and W.~Wang, ``{Expanding Performance Boundaries of Open-Source Multimodal Models with Model, Data, and Test-Time Scaling},'' \emph{arXiv:2412.05271}, 2025.

\bibitem{ye2024mplugowl3longimagesequenceunderstanding}
J.~Ye, H.~Xu, H.~Liu, A.~Hu, M.~Yan, Q.~Qian, J.~Zhang, F.~Huang, and J.~Zhou, ``{mPLUG-Owl3: Towards Long Image-Sequence Understanding in Multi-Modal Large Language Models},'' \emph{arXiv:2408.04840}, 2024.

\bibitem{CLAT}
C.~Wen, M.~Ye, H.~Li, T.~Chen, and X.~Xiao, ``{Concept-Based Lesion Aware Transformer for Interpretable Retinal Disease Diagnosis},'' \emph{IEEE Transactions on Medical Imaging}, vol.~44, no.~1, pp. 57--68, 2025.

\bibitem{silva2025foundation}
J.~Silva-Rodríguez, H.~Chakor, R.~Kobbi, J.~Dolz, and I.~{Ben Ayed}, ``{A Foundation Language-Image Model of the Retina (FLAIR): encoding expert knowledge in text supervision},'' \emph{Medical Image Analysis}, vol.~99, p. 103357, 2025.

\bibitem{codella2024medimageinsight}
N.~C. Codella, Y.~Jin, S.~Jain, Y.~Gu, H.~H. Lee, A.~B. Abacha, A.~Santamaria-Pang, W.~Guyman, N.~Sangani, S.~Zhang \emph{et~al.}, ``{MedImageInsight: An Open-Source Embedding Model for General Domain Medical Imaging},'' \emph{arXiv:2410.06542}, 2024.

\bibitem{he2016deep}
K.~He, X.~Zhang, S.~Ren, and J.~Sun, ``{Deep Residual Learning for Image Recognition},'' in \emph{Proceedings of the IEEE Conference on Computer Vision and Pattern Recognition (CVPR)}, 2016, pp. 770--778.

\bibitem{dosovitskiy2020image}
A.~Dosovitskiy, L.~Beyer, A.~Kolesnikov, D.~Weissenborn, X.~Zhai, T.~Unterthiner, M.~Dehghani, M.~Minderer, G.~Heigold, S.~Gelly, J.~Uszkoreit, and N.~Houlsby, ``{An Image is Worth 16x16 Words: Transformers for Image Recognition at Scale},'' in \emph{International Conference on Learning Representations (ICLR)}, 2021.

\bibitem{hou2024concept}
J.~Hou, J.~Xu, and H.~Chen, ``{Concept-Attention Whitening for Interpretable Skin Lesion Diagnosis},'' in \emph{Medical Image Computing and Computer Assisted Intervention (MICCAI)}, 2024, p. 113–123.

\bibitem{barbano2024ai}
C.~A. Barbano, R.~Renzulli, M.~Grosso, D.~Basile, M.~Busso, and M.~Grangetto, ``{AI-Assisted Diagnosis for Covid-19 CXR Screening: from Data Collection to Clinical Validation},'' in \emph{Proceedings of the IEEE International Symposium on Biomedical Imaging (ISBI)}, 2024, pp. 1--4.

\bibitem{madarapu2024multi}
S.~Madarapu, S.~Ari, and K.~Mahapatra, ``A multi-resolution convolutional attention network for efficient diabetic retinopathy classification,'' \emph{Computers and Electrical Engineering}, vol. 117, p. 109243, 2024.

\bibitem{zhang2023biomedclip}
S.~Zhang, Y.~Xu, N.~Usuyama, H.~Xu, J.~Bagga, R.~Tinn, S.~Preston, R.~Rao, M.~Wei, N.~Valluri, C.~Wong, A.~Tupini, Y.~Wang, M.~Mazzola, S.~Shukla, L.~Liden, J.~Gao, A.~Crabtree, B.~Piening, C.~Bifulco, M.~P. Lungren, T.~Naumann, S.~Wang, and H.~Poon, ``{A Multimodal Biomedical Foundation Model Trained from Fifteen Million Image–Text Pairs},'' \emph{NEJM AI}, vol.~2, no.~1, p. AIoa2400640, 2025.

\bibitem{sun2023eva}
Q.~Sun, Y.~Fang, L.~Wu, X.~Wang, and Y.~Cao, ``{EVA-CLIP: Improved Training Techniques for CLIP at Scale},'' \emph{arXiv:2303.15389}, 2023.

\bibitem{chauhan2023interactive}
K.~Chauhan, R.~Tiwari, J.~Freyberg, P.~Shenoy, and K.~Dvijotham, ``{Interactive Concept Bottleneck Models},'' in \emph{Proceedings of the AAAI Conference on Artificial Intelligence}, vol.~37, no.~5, 2023, pp. 5948--5955.

\bibitem{li2024long}
T.~Li, G.~Zhang, Q.~D. Do, X.~Yue, and W.~Chen, ``{Long-context {LLM}s Struggle with Long In-context Learning},'' \emph{Transactions on Machine Learning Research}, 2025.

\bibitem{zhang2024b}
H.~Zhang, W.~Shao, H.~Liu, Y.~Ma, P.~Luo, Y.~Qiao, N.~Zheng, and K.~Zhang, ``{B-AVIBench: Toward Evaluating the Robustness of Large Vision-Language Model on Black-Box Adversarial Visual-Instructions},'' \emph{IEEE Transactions on Information Forensics and Security}, vol.~20, pp. 1434--1446, 2025.

\bibitem{phung2023best}
M.~Phung, V.~Muralidharan, V.~Rotemberg, R.~A. Novoa, A.~S. Chiou, C.~Y. Sad{\'e}e, B.~Rapaport, K.~Yekrang, J.~Bitz, O.~Gevaert \emph{et~al.}, ``{Best Practices for Clinical Skin Image Acquisition in Translational Artificial Intelligence Research},'' \emph{Journal of Investigative Dermatology}, vol. 143, no.~7, pp. 1127--1132, 2023.

\bibitem{cubdataset}
C.~Wah, S.~Branson, P.~Welinder, P.~Perona, and S.~Belongie, ``{The Caltech-UCSD Birds-200-2011 Dataset},'' 2011.

\end{thebibliography}

\appendix

\section*{Supplementary Material}
{
In this supplementary material, we provide additional details that are not included in the main manuscript. Section \ref{sec:mmices} presents additional ablation studies comparing example selection strategies, namely RICES~\cite{yang2022empirical} and MMICES~\cite{chen2024understanding}. Section \ref{sec:spider_plots} offers further visualizations that illustrate the multidimensional interactions of concepts and model behavior in different LVLMs. Section \ref{sec:implementation_details} includes details of the HuggingFace checkpoints for each adopted Large Vision-Language Model (LVLM) in our experiments. Section \ref{sec:data_preprocessing} provides a list of the clinical concepts for each dataset. Section \ref{sec:prompts} presents the detailed prompts used in our experiments for both concept detection and disease diagnosis tasks, across all datasets.}

\section{{Comparison between example selection strategies - RICES and MMICES}}
\label{sec:mmices}

\begin{figure*}[!h]
    \centering
    \begin{tabular}{c}
        \vspace{-3mm}
        \begin{tikzpicture}
            \node at (0.0,0.0)
            {\includegraphics[width=0.9\textwidth]{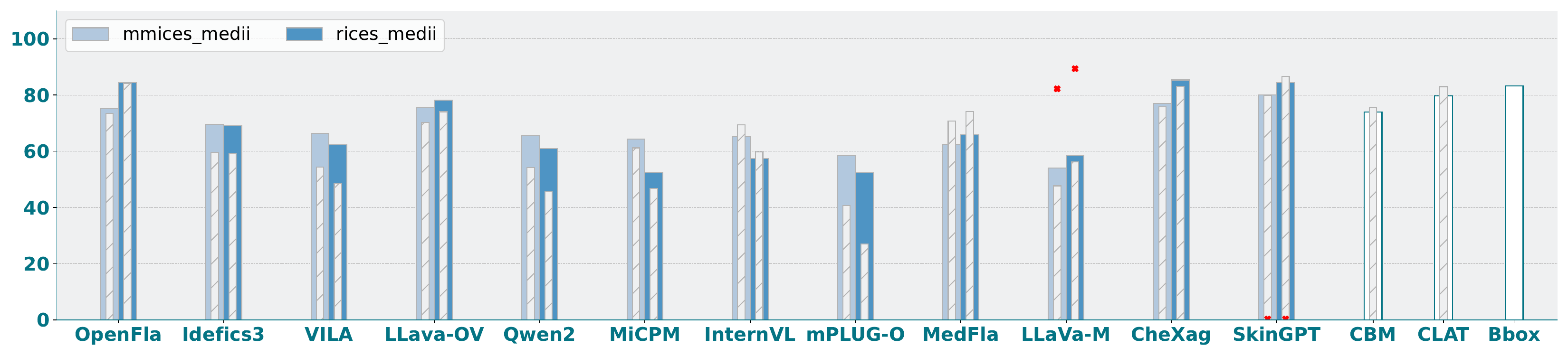}}; 
            \node[font=\footnotesize\sffamily\bfseries, rotate=90, text=black] at (-0.46\textwidth, 0) {Derm7pt};
            \node[font=\footnotesize\sffamily\bfseries, text=gray] at (-0.175\textwidth, 2) {Generic};
            \node[font=\footnotesize\sffamily\bfseries, text=gray] at (0.195\textwidth, 2) {Medical};
            \draw[dashed, thick, gray] (0.071\textwidth, -1.35) -- (0.071\textwidth, 1.645);
            \draw[dashed, thick, gray] (0.315\textwidth, -1.35) -- (0.315\textwidth, 1.645);
        \end{tikzpicture} \\
        \vspace{-3mm}
        \begin{tikzpicture}
            \node at (0, 0) 
            {\includegraphics[width=0.9\textwidth]{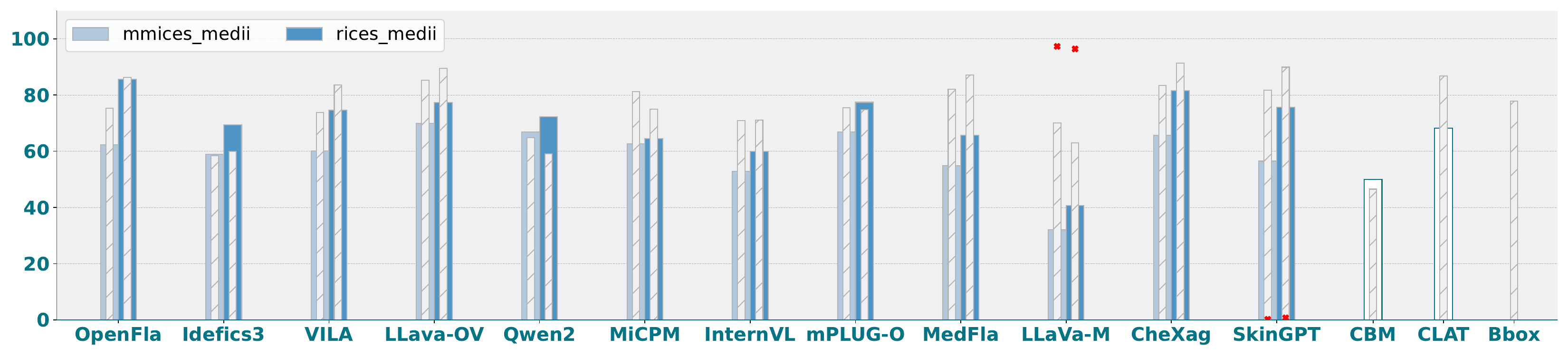}};
            \node[font=\footnotesize\sffamily\bfseries, rotate=90, text=black] at (-0.46\textwidth, 0) {SkinCon};
            \draw[dashed, thick, gray] (0.071\textwidth, -1.35) -- (0.071\textwidth, 1.645);
            \draw[dashed, thick, gray] (0.315\textwidth, -1.35) -- (0.315\textwidth, 1.645);
        \end{tikzpicture} \\
        \vspace{-3mm}
        \begin{tikzpicture}
            \node at (0, 0) 
            {\includegraphics[width=0.9\textwidth]{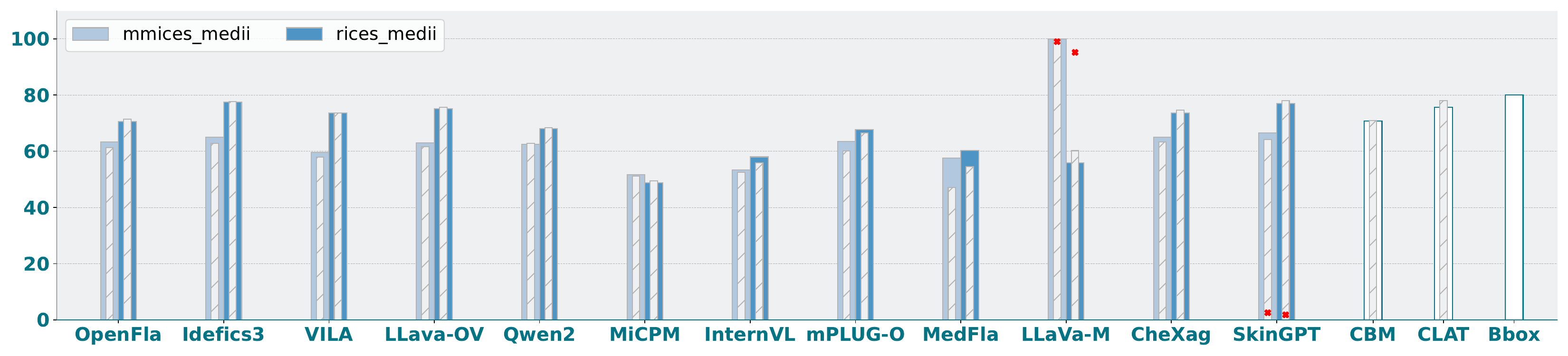}};
            \node[font=\footnotesize\sffamily\bfseries, rotate=90, text=black] at (-0.46\textwidth, 0) {CORDA};
            \draw[dashed, thick, gray] (0.071\textwidth, -1.35) -- (0.071\textwidth, 1.645);
            \draw[dashed, thick, gray] (0.315\textwidth, -1.35) -- (0.315\textwidth, 1.645);
        \end{tikzpicture} \\
        \begin{tikzpicture}
            \node at (0, 0) 
            {\includegraphics[width=0.9\textwidth]{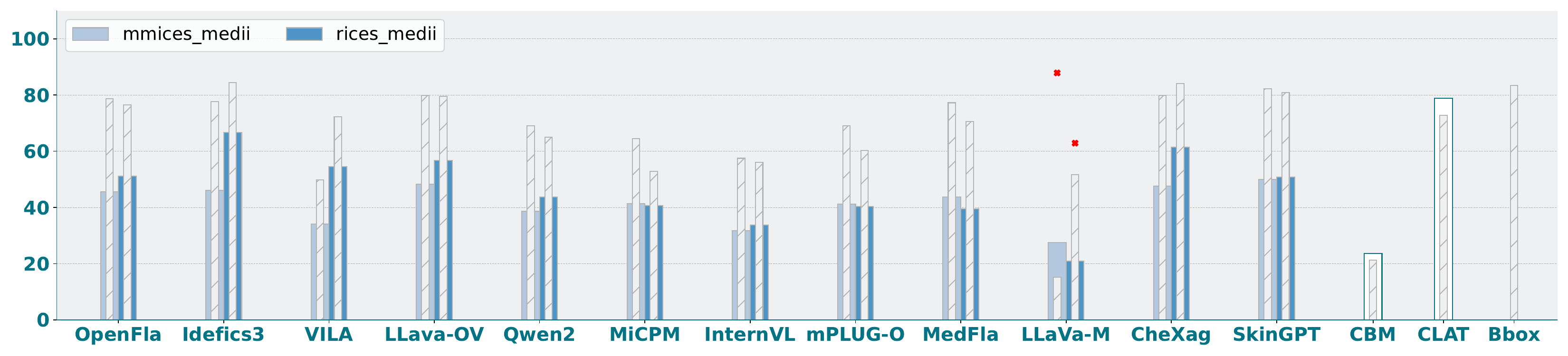}};
            \node[font=\footnotesize\sffamily\bfseries, rotate=90, text=black] at (-0.46\textwidth, 0) {DDR};
            \draw[dashed, thick, gray] (0.071\textwidth, -1.35) -- (0.071\textwidth, 1.645);
            \draw[dashed, thick, gray] (0.315\textwidth, -1.35) -- (0.315\textwidth, 1.645);
        \end{tikzpicture}
    \end{tabular}
    \caption{{\textbf{Disease diagnosis results with 4-shots and two example selection strategies (RICES~\cite{yang2022empirical} and MMICES~\cite{chen2024understanding}).} In both methodologies, MedImageInsight~\cite{codella2024medimageinsight} is the vision encoder used to extract the features from the query and demonstration images.}}
    \label{fig:mmices}
\end{figure*}

{In the main manuscript all experiments are conducted using Retrieval-based In-Context Examples Selection (RICES)~\cite{yang2022empirical} for the selection of the demonstration examples. This method selects the examples for in-context learning (ICL) using the similarity between visual features. Chen et al~\cite{chen2024understanding} extend RICES to also take into account text features, in a method called Mixed Modality In-Context Example Selection (MMICES), which the authors originally tested on vision-language datasets (e.g. image captioning, visual question answering). However, in our datasets such textual features do not exist. Thus, in the first stage of \emph{CBVLM} (concept detection), MMICES cannot be applied. In the second stage (disease diagnosis), MMICES can be applied if we consider the concepts for each image as the textual features; the query image features are first compared to the train dataset image features, of which the $K$ ($K = 200$) most similar are selected. Of these $K$ example images, we select the $n$ most similar to the query according to the cosine similarity between the concept vectors. The results of the comparison between the two approaches are presented in Figure \ref{fig:mmices}, which shows the consistent improvement of RICES over MMICES. This might be explained by the fact that MMICES was proposed for tasks where the textual features have greater inter-image variability (e.g. different image captions for different images), while in our use-cases different images (especially from the same disease) might have the same set of concepts. Furthermore, we use a categorical representation of the set of concepts for each image, not a textual one. In the future it might be interesting to explore how to better adapt MMICES for this scenario or to use an image captioning model to provide textual descriptions of the images.}

\section{{Additional Visualizations}}
\label{sec:spider_plots}

\begin{figure*}[!tb]
    \centering
    \includegraphics[width=\textwidth]{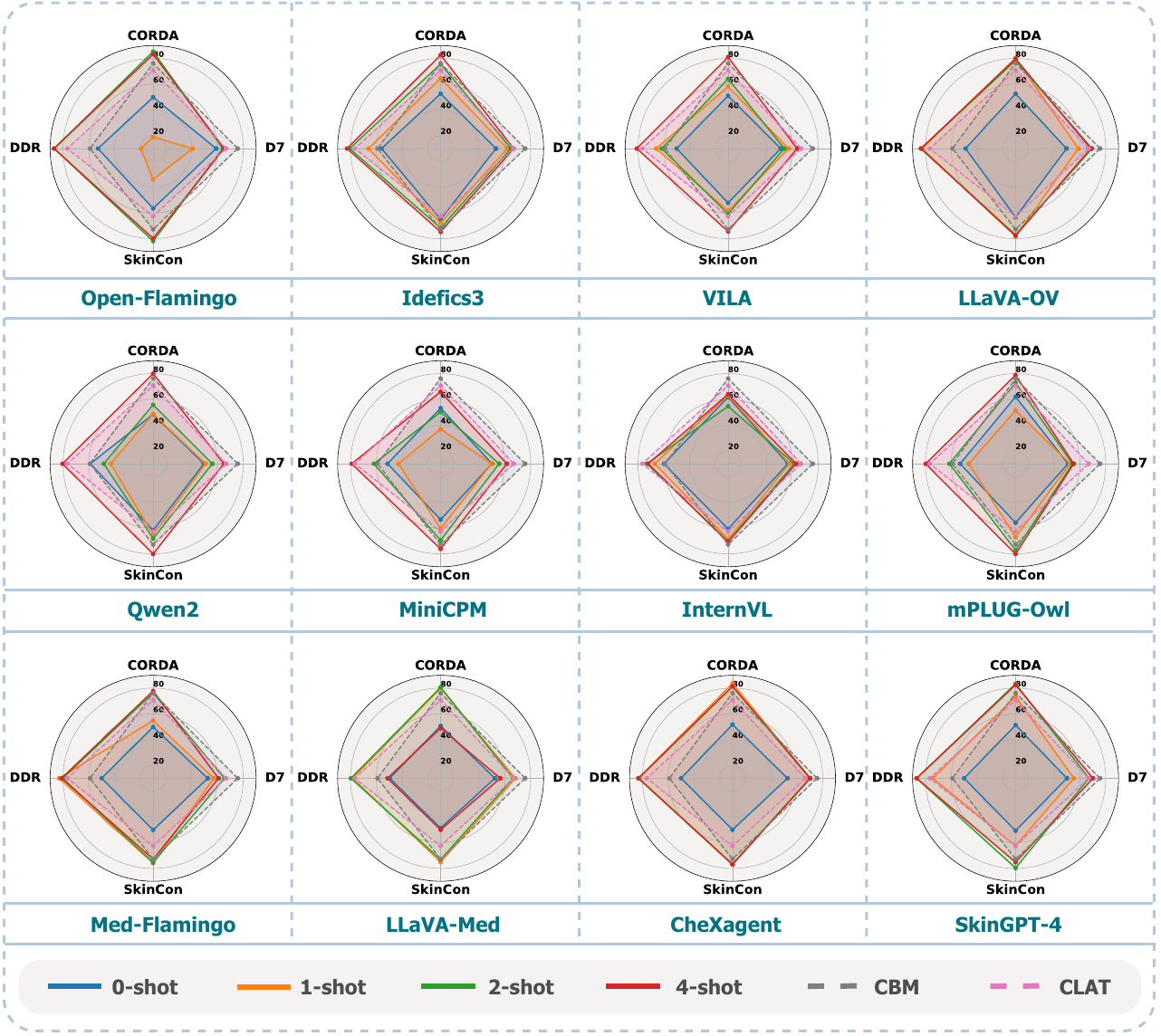}
    \caption{{\textbf{Comparison of LVLM concept detection performance across datasets under different $n$-shot settings, reported in terms of Balanced Accuracy.} Each $n$-shot setting is represented by a different color. The gray and pink dashed lines indicate the performance of CBM and CLAT, respectively.}}
    \label{fig:spider_plots_concept_per_model}
\end{figure*}

\begin{figure*}[!tb]
    \centering
    \includegraphics[width=\textwidth]{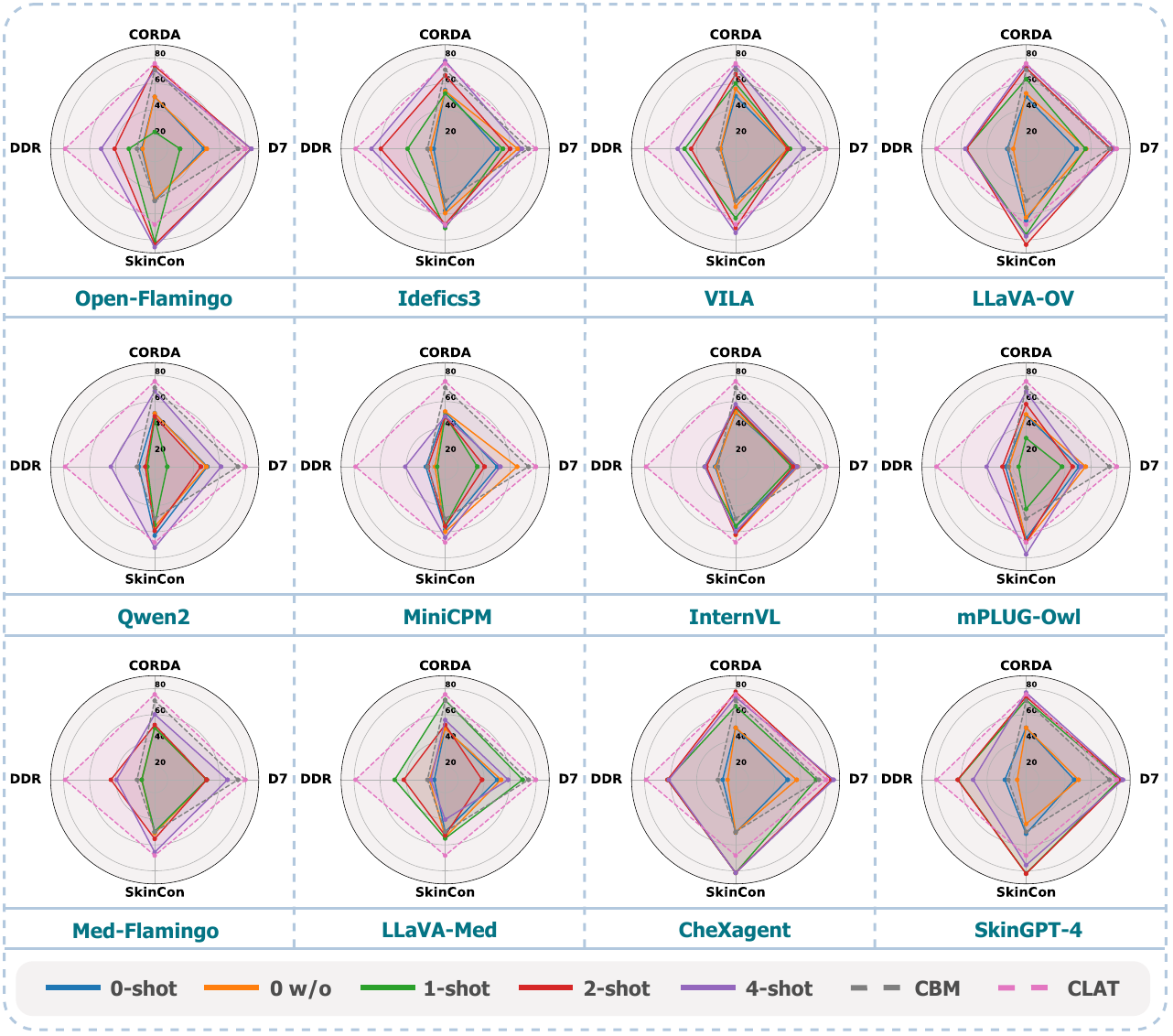}
    \caption{{\textbf{Comparison of LVLM disease diagnosis performance across datasets under different $n$-shot settings, reported in terms of Balanced Accuracy.} Each $n$-shot setting is represented by a different color. The gray and pink dashed lines indicate the performance of CBM and CLAT, respectively.}}
    \label{fig:spider_plots_disease_diagnosis_per_model}
\end{figure*}

{
Figures \ref{fig:spider_plots_concept_per_model} and \ref{fig:spider_plots_disease_diagnosis_per_model} present alternative visualizations to Figures \ref{fig:results_concept_detection} and \ref{fig:results_disease_diagnosis_performance}, respectively, in the form of spider plots, illustrating model behavior across different datasets for the tasks of concept detection and disease diagnosis. It is interesting to note that LVLMs trained on a given domain are able to, without training and with just a few demonstration examples, perform competitively in completely different domains (\textit{c.f.} CheXagent on skin datasets or SkinGPT-4 on CORDA).}

\begin{figure*}[!tb]
    \centering
    \includegraphics[width=\textwidth]{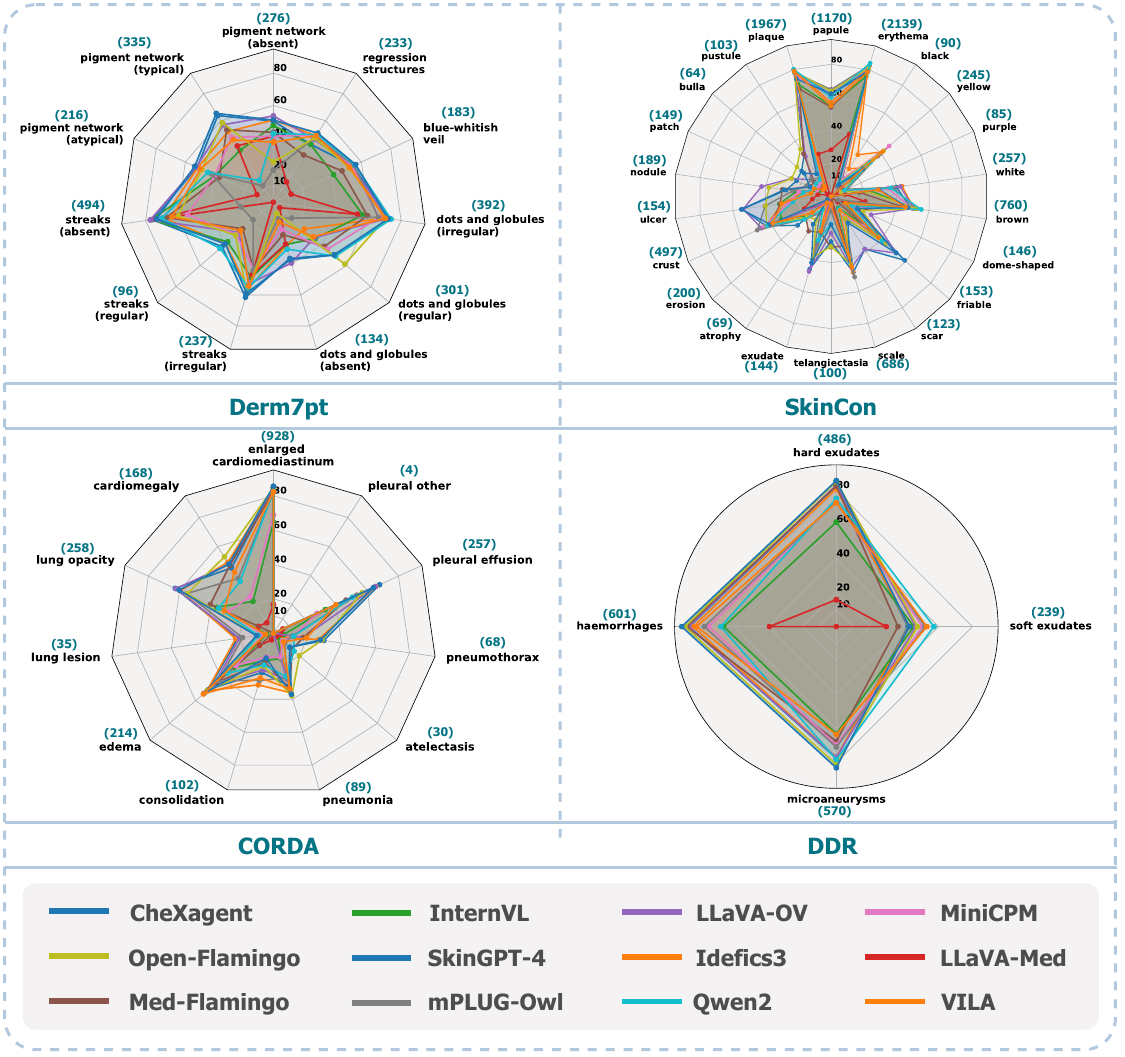}
    \caption{{\textbf{Comparison of per-concept F1-scores across datasets for each evaluated LVLM, illustrating the relative performance patterns of the models in detecting individual clinical concepts.} Values in parentheses indicate the number of samples in the dataset for the corresponding concept.}}
    \label{fig:spider_plots_individual_concept_per_dataset_and_model}
\end{figure*}

{
Figure \ref{fig:spider_plots_individual_concept_per_dataset_and_model} compares the per-concept performance of each evaluated LVLM in each of the four tested datasets, highlighting the LVLMs' ability to detect individual clinical concepts. As mentioned in the main text, and again can be verified here for each LVLM, underrepresented concepts are the ones for which the F1-scores are lower.}

\section{Implementation Details}
\label{sec:implementation_details}

The HuggingFace\footnote{ \url{https://huggingface.co}} checkpoints used for each LVLM are listed in Table \ref{tab:checkpoints}.

For SkinGPT-4~\cite{zhou2024pre}, we used the \texttt{ skingpt4\_llama2\_13bchat\_base\_pre\-train\_stage2} checkpoint available at:
\url{https://github.com/JoshuaChou2018/SkinGPT-4}.

For the automatic extraction of the LVLMs' answers we use the Mistral LLM, specifically the \texttt{mistralai/Mistral-7B-Instruct-v0.3} checkpoint at HuggingFace.

\begin{table}[h]
    \centering
    \caption{HuggingFace checkpoints of the LVLMs used.}
    \setlength{\tabcolsep}{3.5pt}
    \resizebox{\linewidth}{!}{%
    \begin{tabular}{lc}
        \toprule
         \textbf{Model} & \textbf{Checkpoint} \\
         \midrule 
         \midrule
         \multicolumn{2}{c}{\textit{Generic LVLMs}} \\
         \midrule
         OpenFlamingo~\cite{awadalla2023openflamingo} & \texttt{openflamingo/OpenFlamingo-4B-vitl-rpj3b-langinstruct} \\
         {Idefics3}~\cite{idefics3} & \texttt{HuggingFaceM4/Idefics3-8B-Llama3} \\
         VILA~\cite{lin2024vila} & \texttt{Efficient-Large-Model/Llama-3-VILA1.5-8B}\\
         LLaVA-OneVision~\cite{li2024llavaov} & \texttt{lmms-lab/llava-onevision-qwen2-7b-ov} \\
         Qwen2-VL~\cite{Qwen-VL} & \texttt{Qwen/Qwen2-VL-7B-Instruct} \\
         MiniCPM-V~\cite{yao2024minicpm} & \texttt{openbmb/MiniCPM-V-2\_6} \\
         {InternVL 2.5}~\cite{chen2025internvl25} & \texttt{OpenGVLab/InternVL2\_5-8B} \\
         mPLUG-Owl3~\cite{ye2024mplugowl3longimagesequenceunderstanding} & \texttt{mPLUG/mPLUG-Owl3-7B-240728} \\
         \midrule
         \midrule
         \multicolumn{2}{c}{\textit{Medical LVLMs}} \\
         \midrule
         Med-Flamingo~\cite{moor2023med} & \texttt{med-flamingo/med-flamingo} \\
         LLaVA-Med~\cite{li2024llava} & \texttt{microsoft/llava-med-v1.5-mistral-7b} \\
         CheXagent~\cite{chen2024chexagent} & \texttt{StanfordAIMI/CheXagent-8b} \\
         \bottomrule
    \end{tabular}%
    }
    \label{tab:checkpoints}
\end{table}

\section{Clinical Concepts}
\label{sec:data_preprocessing}

Table \ref{tab:dataset_concepts} lists all clinical concepts annotated in each dataset and used in our experiments. All concepts are binary (i.e. are either absent or present) except for ``Pigment Network'', ``Streaks'', and ``Dots and Globules'' of the Derm7pt dataset, for which the options are shown in the table.

\begin{table}[H]
    \centering
    \caption{List of annotated clinical concepts of each dataset.}
    \resizebox{0.5\linewidth}{!}{%
    \begin{tabular}{ll}
        \toprule
        \textbf{Dataset} & \textbf{Clinical Concepts} \\
        \midrule 
        \multirow{5}{*}{Derm7pt} & Pigment Network - \{Absent, Typical, Atypical\}\\
        & Streaks - \{Absent, Regular, Irregular\}\\
        & Dots and Globules - \{Absent, Regular, Irregular\} \\
        & Blue-Whitish Veil \\
        & Regression Structures \\
        \midrule
        \multirow{11}{*}{SkinCon} & Papule, Erythema \\
        & Plaque,  Pustule \\
        & Bulla, Patch \\
        & Nodule, Ulcer \\
        & Crust, Erosion \\
        & Atrophy, Exudate \\
        & Telangiectasia, Scale \\
        & Scar, Friable \\
        & Dome-shaped, Brown(Hyperpigmentation) \\
        & White(Hypopigmentation), Purple \\
        & Yellow, Black \\
        \midrule
        \multirow{8}{*}{CORDA} & Enlarged Cardiomediastinum \\
        & Cardiomegaly \\
        & Lung Opacity \\
        & {Lung Lesion} \\
        & Edema \\
        & Consolidation \\
        & Pneumonia \\
        & {Atelectasis} \\
        & Pneumothorax \\
        & Pleural Effusion \\
        & {Pleural Other} \\
        \midrule
        \multirow{4}{*}{DDR} & Hard exudates \\ 
        & Haemorrhages \\
        & Microaneurysms \\ 
        & Soft exudates \\
        \bottomrule
    \end{tabular}%
    }
    \label{tab:dataset_concepts}
\end{table}

\section{Prompts}
\label{sec:prompts}

In the following section, we provide the prompts used in our experiments.

\paragraph{Concept Detection}
The template prompts used to query LVLMs in the task of concept detection are given in Figures \ref{fig:prompt_derm7pt_concepts} to \ref{fig:prompt_other_datasets_concepts} for each dataset.

\paragraph{Disease Diagnosis}
The template prompts used to query LVLMs in the task of disease diagnosis are given in Figures \ref{fig:prompt_derm7pt_classification} to \ref{fig:prompt_ddr_classification} for each dataset.

\begin{figure*}[t!]
    \centering
    \begin{GrayBox}[\textwidth]
    \scalefont{0.75}
    {{\textit{\textbf{Instruction}}}} \\
    The pigment network consists of intersecting brown lines forming a grid-like reticular pattern. It can be absent, typical, or atypical. A typical pigment network appears as a regular grid-like pattern on dermoscopy, consisting of thin lines that form an even mesh. The spaces (holes) between these lines are relatively uniform in size and shape. In an atypical pigment network the lines forming the network are uneven in thickness, and the holes or spaces between the lines vary in size and shape.\\ \\
    {{\textit{\textbf{Demonstrations}}}} \\
    Consider the following examples: \\ 
    In the \textcolor{promptpurple}{\textless image\textgreater}, the pigment network is:\\
    A) Absent\\
    B) Typical \\
    C) Atypical\\
    Choose one option. Do not provide additional information. \\
    \textcolor{promptblue}{Answer:} A) Absent \\
    (...)
    \\
    \\
    {{\textit{\textbf{Query}}}} \\
    In the \textcolor{promptpurple}{\textless image\textgreater}, the pigment network is:\\
    A) Typical \\
    B) Atypical \\
    C) Absent\\
    Choose one option. Do not provide additional information. \\
    \textcolor{promptblue}{Answer:}
    \end{GrayBox}
    \caption{Example concept detection prompt for the \textbf{Derm7pt} dataset with a multiple-choice concept (``pigment network''). The same template applies to the ``streaks'' and ``dot and globules'' concepts, with the corresponding instruction. The instructions for all Derm7pt concepts can be found in Figure~\ref{fig:prompt_derm7pt_classification}. Only 1 demonstration is shown.}
    \label{fig:prompt_derm7pt_concepts}
\end{figure*}

\begin{figure*}
    \centering
    \begin{GrayBox}[\textwidth]
        \scalefont{0.75}{{\textit{\textbf{Instruction}}}} \\
        The blue-whitish veil appears as an opaque, bluish-white area on the surface of the lesion, giving it a hazy or clouded appearance.\\ \\
        {{\textit{\textbf{Demonstrations}}}} \\
        Consider the following examples: \\ 
        In the \textcolor{promptpurple}{\textless image\textgreater}, the blue-whitish veil is:\\
         A) Present \\
         B) Absent \\
        Choose one option. Do not provide additional information. \\
        \textcolor{promptblue}{Answer:} B) Absent. \\
        (...)
        \\
        \\
        {{\textit{\textbf{Query}}}} \\
        In the \textcolor{promptpurple}{\textless image\textgreater}, the blue-whitish veil is:\\
         A) Present \\
         B) Absent \\
         Choose one option. Do not provide additional information. \\
        \textcolor{promptblue}{Answer:}
    \end{GrayBox}
\caption{Example concept detection prompt for the \textbf{Derm7pt} dataset with a binary concept (``blue-whitish veil''). The same template applies to the ``regression structures'' concept. The instructions for all Derm7pt concepts can be found in Figure~\ref{fig:prompt_derm7pt_classification}. Only 1 demonstration is shown.}
\end{figure*}

\begin{figure*}[t!]
    \centering
    \begin{GrayBox}[\textwidth]
        \scalefont{0.75}{{\textit{\textbf{Instruction}}}} \\
        \{A brief description of the \{concept\}\}.\\ \\
        {{\textit{\textbf{Demonstrations}}}} \\
        Consider the following examples: \\
        In the \textcolor{promptpurple}{\textless image\textgreater}, the \{concept\} is:\\
         A) Present \\
         B) Absent \\
         Choose one option. Do not provide additional information. \\
        \textcolor{promptblue}{Answer:} A) Present. \\
        (...)
        \\ \\
        {{\textit{\textbf{Query}}}} \\
        In the \textcolor{promptpurple}{\textless image\textgreater}, the \{concept\} is:\\
         A) Present \\
         B) Absent \\
         Choose one option. Do not provide additional information. \\
        \textcolor{promptblue}{Answer:}
    \end{GrayBox}
    \caption{Example concept detection prompt for the \textbf{SkinCon}, \textbf{CORDA} and \textbf{DDR} datasets. This template can be applied to each of the 22 SkinCon concepts, {11} CORDA concepts, and 4 DDR concepts (see Table \ref{tab:dataset_concepts}). Figures~\ref{fig:prompt_skincon_classification},~\ref{fig:prompt_corda_classification}, and~\ref{fig:prompt_ddr_classification}, respectively. Only 1 demonstration is shown.}
    \label{fig:prompt_other_datasets_concepts}
\end{figure*}

\begin{figure*}
    \centering
    \begin{GrayBox}[\textwidth]
        \scalefont{0.75}{{\textit{\textbf{Instruction}}}} \\
        Consider the following useful concepts to diagnose melanoma.\\
        The pigment network consists of intersecting brown lines forming a grid-like reticular pattern. It can be absent, typical, or atypical. A typical pigment network appears as a regular grid-like pattern on dermoscopy, consisting of thin lines that form an even mesh. The spaces (holes) between these lines are relatively uniform in size and shape. In an atypical pigment network the lines forming the network are uneven in thickness, and the holes or spaces between the lines vary in size and shape.\\
        Streaks are lineal pigmented projections at the periphery of a melanocytic lesion and include radial streaming (lineal streaks) and pseudopods (bulbous projections). They can be absent, regular, or irregular. Regular streaks are symmetrically arranged around the periphery of the lesion, appearing consistently in both length and spacing. 
        Irregular streaks appear as projections at the periphery of a lesion and are irregular in length, thickness, and distribution.\\ 
        Dots and globules can be absent, regular, or irregular. Regular dots and globules are consistent in size and shape throughout the lesion, appearing either as small, 
        round dots or larger, round globules. Irregular dots and globules vary widely in size and shape, with some appearing small and round, while others are larger and more irregular.\\ 
        The blue-whitish veil appears as an opaque, bluish-white area on the surface of the lesion, giving it a hazy or clouded appearance.\\ 
        Regression structures appear as whitish, scar-like depigmented areas within the lesion, indicating areas where the pigment cells have been destroyed or the lesion has undergone partial regression.\\ \\
        {{\textit{\textbf{Demonstrations}}}} \\
        Consider the following examples: \\ 
        What is the type of skin lesion that is associated with the following dermoscopic concepts: \{concepts\} \\
        Options:\\
        A) Melanoma\\
        B) Nevus\\
        Choose one option. Do not provide additional information.\\ \textcolor{promptblue}{Answer:}
        A) Melanoma. \\
        (...) \\ \\
        {{\textit{\textbf{Query}}}} \\
        What is the type of skin lesion that is associated with the following dermoscopic concepts: \{concepts\} \\
        Options:\\
        A) Melanoma\\
        B) Nevus\\
        Choose one option. Do not provide additional information.\\ \textcolor{promptblue}{Answer:}
    \end{GrayBox}
\caption{Example disease diagnosis prompt for the \textbf{Derm7pt} dataset. Only 1 demonstration is shown.}
\label{fig:prompt_derm7pt_classification}
\end{figure*}

\begin{figure*}
    \centering
    \begin{GrayBox}[\textwidth]
        \scalefont{0.75}{{\textit{\textbf{Instruction}}}} \\
        Consider the following useful concepts to diagnose skin lesions.\\
        A papule is a small, solid, raised lesion with a diameter less than 1 cm, typically palpable and may have a distinct border. \\
        Plaque is a broad, elevated lesion with a diameter greater than 1 cm, often with a rough or scaly surface. \\
        A pustule is a small, elevated lesion containing pus, appearing white or yellowish, often surrounded by erythema. \\
        A bulla is a large, fluid-filled blister with a diameter greater than 1 cm, which can be tense and easily ruptured. \\
        A patch is a flat, discolored area of skin with a diameter greater than 1 cm, often lighter or darker than surrounding skin. \\
        A nodule is a solid, raised lesion deeper than a papule, typically with a diameter greater than 1 cm, and may be palpable. \\
        An ulcer is a break in the skin or mucous membrane with a loss of epidermis and dermis, often presenting as a depressed lesion with irregular borders. \\
        A crust is a dry, rough surface resulting from the drying of exudate or serum on the skin, often forming over an ulcer or wound. \\
        An erosion is a superficial loss of skin that does not extend into the dermis, usually appearing as a moist, depressed area. \\
        Atrophy is the thinning or loss of skin tissue, leading to a depressed appearance, often resulting in a fragile and wrinkled surface. \\
        An exudate is a fluid that oozes out of a lesion or wound, which can be serous, purulent, or hemorrhagic, depending on its composition. \\
        Telangiectasia appears as dilated, small blood vessels near the surface of the skin, often appearing as fine, red or purple lines. \\
        Scale appears as a flake or layer of dead skin cells that may shed from the surface, often seen in conditions like psoriasis or eczema. \\
        A scar is a mark left on the skin after the healing of a wound or injury, which may be flat, raised, or depressed compared to surrounding skin. \\
        Friable skin corresponds to skin that easily breaks or bleeds with minimal trauma, often seen in conditions like malignancies or chronic irritation. \\
        A dome-shaped lesion is a lesion with a rounded, elevated appearance, resembling the shape of a dome, which can be smooth or irregular. \\
        Hyperpigmentation, or an area of the skin that appears brown, corresponds to darkened skin area due to excess melanin production, often seen in conditions like age spots or melasma. \\
        Hypopigmentation, or an area of the skin that appears white, corresponds to lightened skin area due to reduced melanin production, which may result from conditions like vitiligo or post-inflammatory hypopigmentation. \\
        A purple area on the skin might indicate bruising or bleeding beneath the skin, often seen in conditions like purpura or ecchymosis. \\
        A yellow area on the skin corresponds to a color change indicating the presence of bilirubin or lipids, often seen in conditions like jaundice or xanthomas. \\
    \end{GrayBox}
\phantomcaption
\label{fig:prompt_skincon_classification}
\end{figure*}
\begin{figure*}
\ContinuedFloat
    \centering
    \begin{GrayBox}[\textwidth]
        \scalefont{0.75}
        Dark pigmentation often indicates the presence of melanin or necrosis, frequently observed in melanoma or necrotic tissue. \\
        Erythema corresponds to redness of the skin caused by increased blood flow to the capillaries, commonly seen in inflammation or irritation. \\ \\
        {{\textit{\textbf{Demonstrations}}}} \\
        Consider the following examples: \\ 
        What is the diagnosis that is associated with the following concepts: \{concepts\} \\
        Options:\\
        A) Benign\\
        B) Malignant\\
        Choose one option. Do not provide additional information.\\ \textcolor{promptblue}{Answer:}
        A) Benign \\ \\ 
        {{\textit{\textbf{Query}}}} \\
        What is the diagnosis that is associated with the following concepts: \{concepts\} \\
        Options:\\
        A) Benign\\
        B) Malignant\\
        Choose one option. Do not provide additional information. \\ \textcolor{promptblue}{Answer:}
    \end{GrayBox}

\caption{Example disease diagnosis prompt for the \textbf{SkinCon} dataset. Only 1 demonstration is shown.}
\label{fig:prompt_skincon_classification}
\end{figure*}

\begin{figure*}
    \centering
    \begin{GrayBox}[\textwidth] 
    \scalefont{0.75}{{\textit{\textbf{Instruction}}}} \\
        Consider the following useful concepts to diagnose covid-19.\\
        An enlarged cardiomediastinum is an increase in the width of the mediastinum, which may suggest conditions like aortic aneurysm, lymphadenopathy, or heart failure. \\
        Cardiomegaly is an abnormal enlargement of the heart, often indicating underlying conditions such as heart failure, hypertension, or cardiomyopathy. \\
        Lung opacity corresponds to an area on the radiograph where the lung appears more opaque than normal, potentially due to fluid, consolidation, or masses within the lung tissue. \\
        {Lung lesion refers to an abnormal area within the lung that may represent a nodule, mass, or area of infiltrate. It can be benign or malignant and may appear as a localized opacity or irregularity on imaging.} \\
        Edema is the accumulation of fluid in the lung interstitium or alveoli, causing increased opacity on imaging, often indicative of conditions such as heart failure or acute respiratory distress syndrome (ARDS). \\
        Consolidation corresponds to an area of the lung where normal air-filled spaces have been replaced with fluid, cells, or other material, often seen in pneumonia or other infections. \\
        Pneumonia is an infection of the lung parenchyma causing inflammation and consolidation, which appears as areas of opacity on radiographs. \\
        {Atelectasis is the partial or complete collapse of a part of the lung, leading to volume loss and increased opacity on radiographs, which can result from airway obstruction, compression, or surfactant deficiency.} \\
        A pneumothorax corresponds to the presence of air in the pleural space, leading to lung collapse, often visible as a dark area on the radiograph where the lung is not present. \\
        A pleural effusion corresponds to the accumulation of fluid in the pleural space between the lung and chest wall, appearing as a blunting of the costophrenic angles or a meniscus sign on radiographs.\\ 
        {Pleural other refers to pleural abnormalities not classified as effusion or pneumothorax, including conditions such as pleural thickening, plaques, or calcifications.} \\ \\
        {{\textit{\textbf{Demonstrations}}}} \\
        Consider the following examples: \\ 
        What is the diagnosis that is associated with the following concepts: \{concepts\} \\
        Options:\\
        A) COVID-19\\
        B) No COVID-19\\
        Choose one option. Do not provide additional information. \\  \textcolor{promptblue}{Answer:}
        A) COVID-19 \\
        (...)
        \\ \\
        {{\textit{\textbf{Query}}}} \\
        What is the diagnosis that is associated with the following concepts: \{concepts\} \\
        Options:\\
        A) COVID-19\\
        B) No COVID-19\\
        Choose one option. Do not provide additional information.\\ \textcolor{promptblue}{Answer:}
    \end{GrayBox}
\caption{Example disease diagnosis prompt for the \textbf{CORDA} dataset. Only 1 demonstration is shown.}
\label{fig:prompt_corda_classification}
\end{figure*}

\begin{figure*}[tb!]
    \centering
    \begin{GrayBox}[0.95\textwidth]
        \scalefont{0.75}{{\textit{\textbf{Instruction}}}} \\
        Consider the following useful concepts to diagnose diabetic retinopathy.\\
        Hard exudates typically appear as yellowish or whitish, well-defined spots or patches on the retina. They can vary in size and shape and often form clusters or rings around areas of retinal edema (swelling).\\
        Soft exudates, also known as cotton-wool spots appear as pale, fluffy, or cloud-like spots on the retina. Unlike hard exudates, they are not well-defined and have a more feathery edge.\\
        Haemorrhages can appear as red or dark spots, streaks, or blotches on the retina.\\
        Microaneurysms appear as tiny, red dots on the retina. They are typically round and uniform in shape. They can be distinguished from other retinal features by their small size and bright red color.\\ \\
        {{\textit{\textbf{Demonstrations}}}} \\
        Consider the following examples:\\
        What is the type of diabetic retinopathy that is associated with the following concepts: \{concepts\} \\
        Options:\\
        A) No diabetic retinopathy \\
        B) Mild diabetic retinopathy \\
        C) Moderate diabetic retinopathy \\
        D) Severe diabetic retinopathy \\
        E) Proliferative diabetic retinopathy \\
        Choose one option. Do not provide additional information.\\
        \textcolor{promptblue}{Answer:}
        B) Mild diabetic retinopathy \\
        (...) \\ \\
        {{\textit{\textbf{Query}}}} \\
        What is the type of diabetic retinopathy that is associated with the following concepts: \{concepts\} \\
        Options:\\
        A) No diabetic retinopathy \\
        B) Mild diabetic retinopathy \\
        C) Moderate diabetic retinopathy \\
        D) Severe diabetic retinopathy \\
        E) Proliferative diabetic retinopathy \\
        Choose one option. Do not provide additional information.\\ \textcolor{promptblue}{Answer:}
    \end{GrayBox}
\caption{Example disease diagnosis prompt for the \textbf{DDR} dataset. Only 1 demonstration is shown.}
\label{fig:prompt_ddr_classification}
\end{figure*}

\paragraph{Automatic Evaluation.}
The template prompt used to query Mistral to perform the automatic extraction of the LVLMs' responses is shown in Figure \ref{fig:prompt_mistral}.

\begin{figure*}[h!]
    \centering
    \begin{GrayBox}[0.95\textwidth]
        \scalefont{0.8}
        Sentence: \textless\textless\textless\{sentence\}\textgreater\textgreater\textgreater\\
        Consider the sentence given in between \textless\textless\textless \textgreater\textgreater\textgreater\ and the following options: \{options\}\\
        Choose the option that best fits the information conveyed by the sentence. 
        Take your time and answer in json format by providing only the letter corresponding to the chosen option, following the template: \{\{`Answer': `option letter'\}\}. 
        If the sentence does not provide enough information to choose an option, provide the following answer: \{\{`Answer': `UNK'\}\}.\\
        Do not provide additional responses, context or explanations.\\
        \textcolor{promptblue}{Answer:}
    \end{GrayBox}
\caption{Prompt to perform automatic extraction of the LVLMs' responses with the \textbf{Mistral} LLM.}
\label{fig:prompt_mistral}
\end{figure*}

\end{document}